\documentclass[applsci,article,accept,moreauthors]{Definitions/mdpi}

\firstpage{1}
\pubvolume{16}
\issuenum{12}
\articlenumber{6153}
\pubyear{2026}
\copyrightyear{2026}
\datereceived{7 May 2026}
\daterevised{13 June 2026}
\dateaccepted{14 June 2026}
\datepublished{17 June 2026}

\usepackage{algorithm}
\usepackage{algorithmic}
\renewcommand{\textcolor}[2]{#2}
\externalbibliography{yes}
\Title{Dynamic LoRA-Experts and Prototype-Ensemble Matching for Class-Incremental Learning}

\Author{Hongwei Zhao $^{1}$*, Rui Liu $^{1}$ and Yansong Liu $^{1}$}

\AuthorNames{Hongwei Zhao, Rui Liu and Yansong Liu}

\address{%
$^{1}$ \quad School of Computer Science and Engineering, Beihang University, 37 Xueyuan Road, Beijing 100191, China}

\corres{Correspondence: zhaohongwei@buaa.edu.cn (H.Z.)}

\abstract{Class-Incremental Learning (CIL) aims to continuously learn new classes without forgetting previously acquired knowledge. Recent advances in parameter-efficient fine-tuning (PEFT) based on pre-trained models (PTMs) have shown promise in this setting by integrating new tasks with minimal parameter overhead. However, these methods often suffer from \textit{knowledge degradation} due to: (1) cumulative interference caused by iterative updates, constrained gradient flows, or entangled module integration; and (2) suboptimal alignment between inference samples and specialized modules. \textcolor{red}{To address these challenges, we propose Dynamic LoRA-Experts and Prototype-Ensemble Matching (DLEPEM), a novel two-stage, rehearsal-free framework.} In the first stage, we allocate a task-specific LoRA-Expert for each incremental task, enabling isolated representation learning and reducing cross-task interference. In the second stage, we introduce a prototype-ensemble matching mechanism that combines general prototypes derived from the frozen PTM with task-adaptive prototypes learned by the LoRA-Experts. This fusion facilitates both strong generalization and precise task-level discrimination. 
\textcolor{red}{Extensive experiments on standard CIL and Few-Shot Class-Incremental Learning (FSCIL) benchmarks demonstrate that DLEPEM achieves strong performance under the evaluated protocols. For instance, in CIL, it achieves \textbf{93.39\%} on CIFAR100 (+0.80\% over EASE), \textbf{92.31\%} on CUB200 (+2.11\% over EASE), and \textbf{91.84\%} on VTAB (+1.39\% over EASE). In the more challenging FSCIL setting, it achieves \textbf{88.77\%} on CUB200, outperforming the strongest baseline by a clear margin of \textbf{5.31\%}. These results indicate that DLEPEM effectively mitigates catastrophic forgetting while enhancing incremental learning capability. Code is available at: \url{https://github.com/hongwei-zhao/Applied_Sciences-DLEPEM-main}.}}


\featuredapplication{This work can be applied to intelligent vision systems that need to incrementally learn new categories while maintaining previously acquired knowledge with low storage overhead.}

\begin{document}

\section{Introduction}
In open-world environments, data typically arrives as a continuous stream of novel categories, a scenario formalized as Class-Incremental Learning (CIL). Traditional machine learning models struggle under such conditions, exhibiting catastrophic forgetting~\cite{mccloskey1989catastrophic,french1999catastrophic}, where learning from new classes disrupts existing representations and leads to severe performance degradation. Class-incremental learning seeks to navigate this challenge by balancing the acquisition of new knowledge with the preservation of prior learning, a fundamental trade-off known as the stability-plasticity dilemma~\cite{grossberg2012studies,liang2024inflora}.

Recent advances leverage pre-trained models (PTMs), whose robust generalization capabilities stem from large-scale supervised or self-supervised training~\cite{han2021pre}, providing a compelling foundation for CIL. However, directly fine-tuning all PTM parameters across sequential tasks risks compromising this generalization and amplifying forgetting. To mitigate this, contemporary methods use parameter-efficient fine-tuning (PEFT) techniques~\cite{xin2024parameter}, such as \textit{prompts}~\cite{wang2022learning,wang2022dualprompt}, \textit{adapters}~\cite{gao2023unified,zhou2023revisitingclassincrementallearningpretrained}, and \textit{LoRA}~\cite{liang2024inflora,wu2025sdlora}, which adapt models with minimal additional parameters, thereby reducing forgetting while preserving generalization.

Despite these gains, two critical challenges remain insufficiently addressed:

\textbf{1. Stability-plasticity limitations from cumulative interference}:
\begin{itemize}
    \item 
    Shared prompt pools~\cite{wang2022learning,wang2022dualprompt} are prone to overwriting earlier knowledge when exposed to shifting data distributions.
    
    \item 
    LoRA-based strategies~\cite{liang2024inflora,wu2025sdlora}, though effective in constraining updates to mitigate forgetting, inadvertently restrict plasticity needed for new task adaptation.
    
    \item 
    Fusion-based methods~\cite{gao2023unified,liang2024inflora} attempt to balance old and new knowledge but often degrade the fidelity of both due to forced trade-offs.
\end{itemize}

\textbf{2. Inference-stage module-sample mismatches}:
\begin{itemize}
    \item 
    Fixed PTM selection mechanisms~\cite{wang2022learning,wang2022s} struggle under substantial domain shifts, resulting in suboptimal activations and degraded predictions.
\end{itemize}
%
These observations motivate a key question: Can we simultaneously enhance stability-plasticity dynamics and improve module-sample matches to robustly mitigate catastrophic forgetting in CIL?

To this end, we propose \textbf{DLEPEM}, a novel rehearsal-free framework that rethinks PEFT-based CIL by introducing two synergistic components. 
First, inspired by Mixture-of-Experts (MoE) architectures~\cite{6797059}, we dynamically allocate a dedicated LoRA-Expert for each new incremental task. \textcolor{red}{Unlike shared or sequentially fine-tuned modules, each LoRA-Expert exclusively encodes task-specific knowledge; only the current expert is trainable, while all prior experts are frozen.} Strategically embedded within Transformer feed-forward network (FFN) and multi-head self-attention (MHA) layers, these lightweight experts preserve plasticity for new tasks while entirely isolating past parameters, thereby achieving a more favorable balance between stability and plasticity. \textcolor{red}{Different from conventional token-level sparse MoE or LoRA-MoE models that learn a soft router and aggregate multiple expert outputs, DLEPEM uses MoE as a task-level organization principle: the expert bank grows with the task sequence, each old expert remains frozen, and expert activation is determined by prototype retrieval rather than differentiable token routing.}
Second, to mitigate inference mismatches, we introduce a prototype ensemble strategy that jointly leverages (1) general representations from the frozen PTM backbone and (2) specialized representations from task-specific experts. By fusing these distinct feature spaces, our approach enhances sample-to-module alignment, effectively bridging generalization and specialization. \textcolor{red}{Thus, DLEPEM explicitly separates within-task prediction (WTP), handled by isolated task LoRA-Experts, from module-identity inference (MII), handled by prototype-ensemble matching.}

In summary, our principal contributions are threefold:
\begin{itemize}
    \item 
    \textcolor{red}{We introduce task-level dynamic LoRA-Expert allocation into PEFT-based CIL. Unlike conventional MoE-style LoRA methods with a fixed expert pool and soft expert mixing, DLEPEM automatically adds one dedicated LoRA-Expert for each incremental task, trains only the current expert, and freezes all previous experts to reduce cross-task parameter interference.}

    \item 
    \textcolor{red}{We enhance module-sample alignment through prototype-ensemble matching, which fuses frozen-PTM prototypes and router-based domain-specific prototypes. This design improves task-level expert retrieval when PTM-only matching is unreliable under downstream domain shift.}

    \item 
    Extensive experiments on six challenging CIL benchmarks validate the effectiveness of our approach, showing that DLEPEM achieves leading performance among the evaluated methods under the evaluated protocols. We further demonstrate architectural flexibility through DLEPEM-MLP, a variant that explores alternative expert integration strategies while retaining competitive results.
\end{itemize}

\section{Related Work}
\subsection{Class-Incremental Learning}
\textcolor{red}{Class-incremental learning requires models to continually recognize newly introduced classes while maintaining discriminability for previously learned classes. Existing methods are commonly grouped into regularization-, rehearsal-, and architecture-based approaches~\cite{wang2022learning}.}

\textcolor{red}{Regularization-based methods~\cite{aljundi2019task} reduce forgetting by penalizing changes to parameters that are estimated to be important for old tasks. This strategy does not store old samples, but its effectiveness can decrease when the incremental stream contains large domain shifts or many sequential tasks~\cite{rebuffi2017icarl}. Rehearsal-based methods replay raw images~\cite{rebuffi2017icarl} or stored feature representations~\cite{yu2020semantic} to preserve old knowledge. They are often effective, but the buffer budget and possible privacy restrictions limit their applicability in rehearsal-free settings. Dynamic network methods~\cite{wang2022foster} expand the model with task-specific components and freeze previous ones to reduce parameter interference, but the growing architecture may introduce non-trivial memory overhead and some methods still rely on old data for calibration or fusion.}

\subsection{PEFT-Based CIL}
\textcolor{red}{With the strong transferability of PTMs, recent CIL methods increasingly adopt PEFT modules to adapt only a small subset of parameters while keeping most backbone weights frozen. Prompt-based methods, such as L2P~\cite{wang2022learning}, DualPrompt~\cite{wang2022dualprompt}, S-Prompts~\cite{wang2022s}, and CODA-Prompt~\cite{smith2023coda}, learn task-relevant prompts and retrieve or combine them during inference. These methods reduce the need for full fine-tuning, but their retrieval quality depends heavily on how well prompt keys separate tasks in the PTM feature space.}

\textcolor{red}{Adapter- and LoRA-based methods modify the PTM through lightweight modules. APER~\cite{zhou2023revisitingclassincrementallearningpretrained} combines PEFT-adapted and PTM features to retain generalization. LAE~\cite{gao2023unified} improves compatibility across PEFT modules, but repeated feature fusion can introduce stability-plasticity trade-offs. InfLoRA~\cite{liang2024inflora} constrains LoRA updates through gradient-orthogonal projection to reduce interference, whereas SD-LoRA~\cite{wu2025sdlora} decouples gradient direction and magnitude to protect early-task directions. These constraint-based strategies improve stability, but may restrict plasticity for newly introduced classes. The MoE-Adapters method~\cite{MoE-Adapters} uses an activate-freeze mechanism with a predefined expert pool, which enables inter-task collaboration but limits flexibility when the number or diversity of future tasks is unknown.}

\subsection{Mixture-of-Experts and Expert Retrieval}
\textcolor{red}{MoE architectures introduce multiple expert modules and a routing mechanism that selects or weights experts for each input~\cite{6797059}. In vision models, V-MoE~\cite{riquelme2021scaling} replaces part of the dense feed-forward layers in ViT with sparse expert layers, where image patches are routed to a subset of MLP experts. In multi-modal or instruction-tuning settings, MoCLE~\cite{gou2023mixture} integrates multiple LoRA experts to handle task diversity. These methods show that expert specialization can improve adaptation, but their routing is usually learned within a fixed or predefined expert set and often combines expert outputs through token-level or sample-level gating.}

\textcolor{red}{For incremental learning, the key challenge is different: the model must accommodate an open-ended task sequence while avoiding repeated updates to old task-specific parameters. Therefore, expert life cycle and inference-time expert retrieval become central design choices. A fixed expert pool may be insufficient for long or unpredictable task streams, while a learned soft gate can suffer from task-recency bias when trained only on the current task.}

\subsection{Our Approach}
\textcolor{red}{Similar to PEFT-based CIL methods~\cite{gao2023unified,zhou2023revisitingclassincrementallearningpretrained,liang2024inflora}, DLEPEM uses a PTM backbone with LoRA~\cite{hu2022lora} adaptation. However, instead of repeatedly updating or blending shared PEFT modules, DLEPEM allocates one LoRA-Expert for each incremental task, trains only the current expert, and freezes all historical experts. This stage-wise expert life cycle reduces cross-task parameter interference while preserving plasticity for the new task.}

\textcolor{red}{DLEPEM also differs from conventional MoE-based or multi-LoRA methods in routing granularity. Existing MoE formulations usually learn an input-dependent gate to select or softly combine experts from a predefined pool. In contrast, DLEPEM performs top-1 task-level expert retrieval through an append-only prototype dictionary. Each ensemble key combines a frozen-PTM prototype with a router-domain prototype, so inference uses both general semantics and task-adaptive cues to select the appropriate expert. Thus, old-task preservation is mainly achieved by structural parameter isolation, while module-identity inference is handled by prototype-ensemble matching rather than a learned soft gate.}

\section{Preliminaries}
\textbf{Problem Formulation.} CIL considers a sequential stream of tasks $\mathcal{D} = \{\mathcal{D}_1, \dots, \mathcal{D}_T\}$, where the $t$-th task $\mathcal{D}_t = \{(\mathbf{x}_i, \mathbf{y}_i)\}_{i=1}^{n_t}$ contains $n_t$ samples. Here, $\mathbf{x}_i \in \mathcal{X}_t$ denotes an input from domain $\mathcal{X}_t$, and $\mathbf{y}_i \in \mathcal{Y}_t$ is its corresponding label. Importantly, the label spaces are mutually exclusive across tasks, i.e., $\mathcal{Y}_t \cap \mathcal{Y}_{t'} = \emptyset$ for $t \neq t'$.

Following the rehearsal-free setting~\cite{wang2022learning,wang2022dualprompt,smith2023coda}, the model only observes data from the current task during training. The training objective is to learn a model $\text{f}_{\Theta}(\mathbf{x}) = \mathbf{W}_{cls}^\top \phi(\mathbf{x})$ that minimizes the empirical risk over the current task's training set:
\begin{equation}
\text{L}(\mathcal{D}_t) = \frac{1}{{|\mathcal{{D}}_t|}}\sum_{(\mathbf{x}_i, \mathbf{y}_i) \in \mathcal{D}_t} \text{L}\bigl(\text{f}_{\Theta}(\mathbf{x}_i), \mathbf{y}_i\bigr),
\label{eq:cilrisk}
\end{equation}
where $\phi(\mathbf{x})$ represents the embedded [class] token from the ViT, $\mathbf{W}_{cls}$ denotes the classifier weights, ${|\mathcal{{D}}_t|}$ is the number of examples in the current task, and $\text{L}(\cdot,\cdot)$ represents the loss function that measures prediction error. After each task $t$, performance is evaluated on all classes seen so far, i.e., on the union $\mathbf{Y}_t = \mathcal{Y}_1 \cup \cdots \cup \mathcal{Y}_t$.

\textcolor{red}{For clarity, Table~\ref{tab:notation} summarizes the main symbols used throughout the paper.}

\textbf{Mixture of Experts.} MoE architectures offer an efficient way to increase model capacity by activating only a subset of parameters per input, achieving faster training and inference compared to dense networks of equivalent scale. A typical MoE layer comprises a set of $M$ expert networks $\mathbf{E} = \{\mathbf{E}_1, \dots, \mathbf{E}_M\}$ and a router $\text{G}$ that determines expert activations based on the input~\cite{jin2025moe}. Each expert is commonly implemented as an FFN, while the router is parameterized by a weight matrix $\mathbf{W}_g$.

Formally, for an input $\mathbf{x}$, the router computes:
\begin{equation}
\text{G}(\mathbf{x}) = \text{softmax}(\mathbf{W}_g \mathbf{x}),
\label{eq:moe-router}
\end{equation}
producing a soft selection over experts. The final MoE output is given by:
\begin{equation}
\text{MoE}(\mathbf{x}) = \sum_{i=1}^M \text{G}(\mathbf{x})_i \mathbf{E}_i(\mathbf{x}),
\label{eq:moe-output}
\end{equation}
where $\text{G}(\mathbf{x})_i$ represents the routing probability for expert $\mathbf{E}_i$. In Transformer-based architectures, MoE layers often replace standard FFN blocks to selectively route representations~\cite{dou2023loramoe}.

\textbf{Low-Rank Adaptation.} LoRA was introduced to efficiently fine-tune large pre-trained models by injecting low-rank updates into weight matrices~\cite{hu2022lora}. Given a pre-trained weight matrix $\mathbf{W} \in \mathbf{R}^{d_{in} \times d_{out}}$, LoRA learns an additive low-rank decomposition:
\begin{equation}
\mathbf{W} + \Delta \mathbf{W} = \mathbf{W} + \mathbf{U} \mathbf{V},
\label{eq:lora}
\end{equation}
where $\mathbf{U} \in \mathbf{R}^{d_{in} \times r}$, $\mathbf{V} \in \mathbf{R}^{r \times d_{out}}$, and the rank $r \ll \min(d_{in}, d_{out})$. This reduces the number of trainable parameters while maintaining expressiveness. LoRA enables cost-effective, scalable fine-tuning, making it suitable for continual learning, where efficiency and avoidance of forgetting are critical.

\section{The Proposed Method}
\begin{figure*}
    \centering
    \includegraphics[width=0.99\linewidth,height=0.27\textheight]{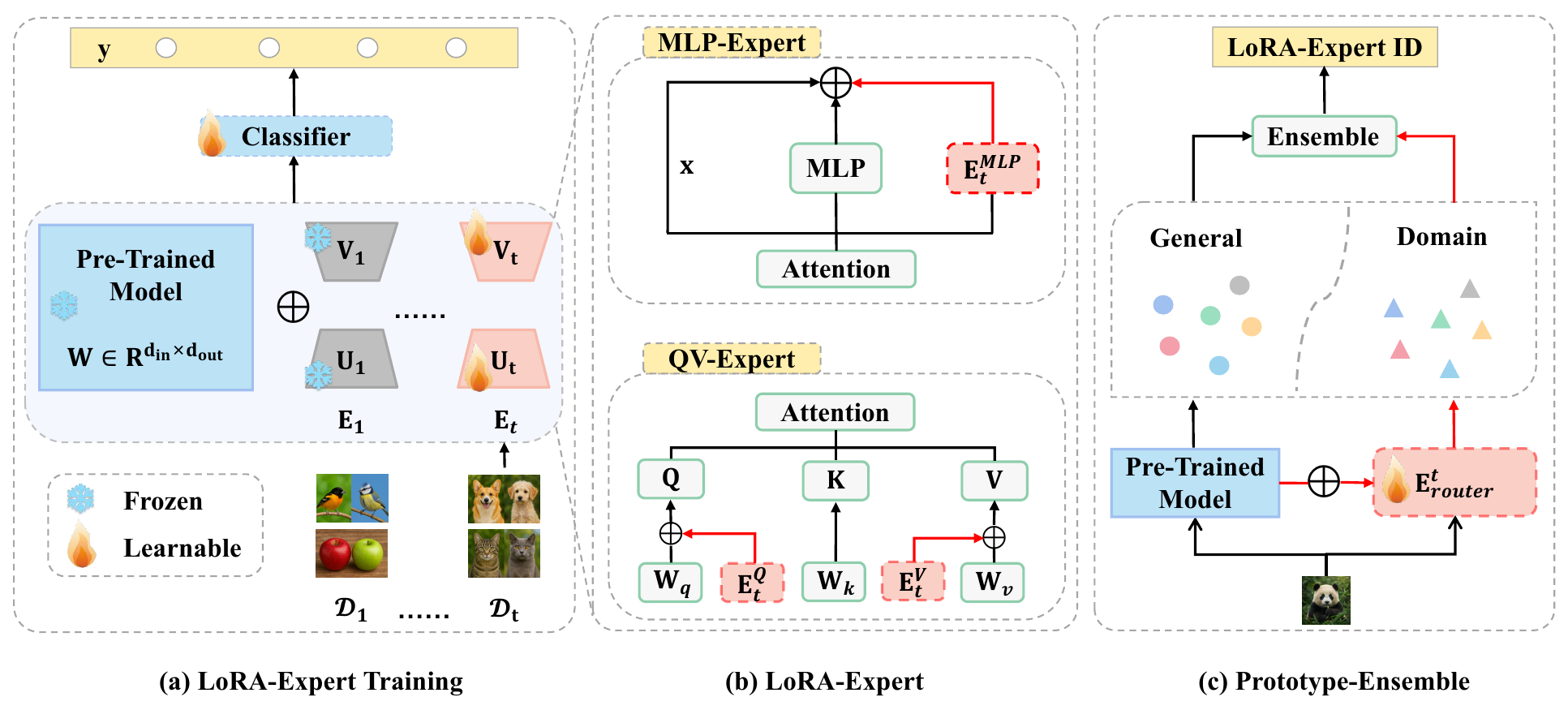}
     \caption{ Illustration of DLEPEM. (a) In the $t$-th incremental task, a new LoRA-Expert $\mathbf{E}_t$ (with parameters $\mathbf{U}_t$ and $\mathbf{V}_t$) is trained to capture task-specific features. (b) Structure of the LoRA-Expert. Depending on the insertion location, the module is categorized as either an MLP-Expert or a QV-Expert. The QV-Expert integrates LoRA into the $\mathbf{W}_q$ and $\mathbf{W}_v$ projections of the MHA layer. (c) Prototype-Ensemble Matching. General prototypes from the PTM and domain-specific prototypes from the router $\mathbf{E}^{router}_{t}$ are combined and linked to their respective experts. During inference, the nearest prototype guides expert selection for each input sample.
    }
    \label{DLEPEM}
\end{figure*}

As demonstrated by HiDe-Prompt~\cite{wang2023hierarchical}, CIL methods employing multi-module selection can be decomposed into two probabilistic components: \textbf{module-identity inference (MII)} and \textbf{within-task prediction (WTP)}, represented by
${P}(\mathbf{x} \in \mathcal{X}_{i}|\mathcal{D},\Theta)$ and
${P}(\mathbf{x} \in \mathcal{X}_{i,j}|\mathbf{x} \in \mathcal{X}_{i},\mathcal{D},\Theta)$, respectively. By \textit{Bayes' theorem}, we have
\begin{align}\label{BayesTheorem}
   {P}(\mathbf{x} \in \mathcal{X}_{i,j}|\mathcal{D},\Theta) = {P}(\mathbf{x} \in \mathcal{X}_{i,j}|\mathbf{x} \in \mathcal{X}_{i},\mathcal{D},\Theta) P(\mathbf{x} \in \mathcal{X}_{i}|\mathcal{D},\Theta).
\end{align}
%
Letting $\hat{i}$ and $\hat{j}$ denote the ground-truth task index and class label for input $\mathbf{x}$, Eq.~\eqref{BayesTheorem} implies that improving either WTP accuracy,
$P(\mathbf{x} \in \mathcal{X}_{\hat{i},\hat{j}}|\mathbf{x} \in \mathcal{X}_{\hat{i}},\mathcal{D},\Theta)$,
or MII accuracy,
$P(\mathbf{x} \in \mathcal{X}_{\hat{i}}|\mathcal{D},\Theta)$,
directly enhances overall prediction performance.
However, existing approaches suffer from two limitations: (1) iterative updates or fusion of new and existing modules progressively deteriorate WTP~\cite{wang2022learning,wang2022dualprompt,gao2023unified,liang2024inflora}; and (2) MII performance, when relying solely on pre-trained features~\cite{wang2022s,gao2023unified}, is inherently constrained by the similarity between pre-training and downstream data distributions.
To address these challenges, we propose \textbf{DLEPEM}, which explicitly enhances both WTP and MII via two complementary innovations:

\paragraph{Dynamic LoRA-Expert} 
To exploit the strong generalization of the pre-trained model, we keep its weights $\mathbf{W}$ fixed throughout training. To maintain plasticity and safeguard WTP, we dynamically allocate a dedicated LoRA-Expert for each incremental task, embedded within an MoE framework. Each new LoRA-Expert is trained exclusively on its respective task while previously introduced experts remain frozen. This ensures isolated task-specific adaptation with a small number of trainable parameters, leveraging LoRA's efficiency to effectively capture discriminative features. Figure~\ref{DLEPEM}(a) depicts expert training, while Figure~\ref{DLEPEM}(b) details the internal structure. 
\textcolor{red}{This expert life cycle is the main difference from conventional MoE-based LoRA continual learning. DLEPEM does not repeatedly update a shared expert pool or combine all experts through a soft gate during feature computation. Instead, for task $t$, only $\mathbf{E}_{t}$ receives gradients, whereas $\{\mathbf{E}_{1},\ldots,\mathbf{E}_{t-1}\}$ remain frozen. This design directly reduces cross-task parameter interference and protects WTP for previously learned tasks.}

\paragraph{Prototype-Ensemble Matching}
After training each incremental task, we extract prototypes using two sources: (1) the fixed PTM for generalizable features, and (2) the router-enhanced LoRA-Expert for domain-specific nuances. \textcolor{red}{Unlike prior methods that rely solely on pre-trained representations, DLEPEM combines these prototypes and associates them with their corresponding LoRA-Experts.} During inference, a nearest-neighbor search in this combined prototype space determines the most appropriate expert, substantially improving MII. This design mitigates privacy concerns inherent to rehearsal-based strategies. Figure~\ref{DLEPEM}(c) illustrates our complete prototype-ensemble matching mechanism.
\textcolor{red}{The ensemble key for a class is constructed as the concatenation of a frozen-PTM prototype and a router-domain prototype. The former provides stable general semantics, while the latter captures downstream task-specific cues. This complements dynamic expert allocation: the task expert improves WTP once selected, and the prototype-ensemble dictionary improves MII by selecting the appropriate expert without storing raw rehearsal samples.}

\subsection{Dynamic LoRA-Experts} 
We integrate the proposed LoRA-Expert modules into the Vision Transformer (ViT) architecture~\cite{dosovitskiy2020image}. In ViT, an input image is first divided into fixed-size patches, linearly projected, and augmented with positional embeddings before being processed by a Transformer encoder comprising MHA layers and multilayer perceptrons (MLPs).

\textcolor{red}{To adaptively capture task-specific features in incremental learning, we dynamically introduce a LoRA-Expert at each incremental stage. This module can be inserted either as a parallel branch to the MLP (MLP-Expert, see Figure~\ref{DLEPEM}(b)) or into the attention projections of the MHA module. When LoRA is applied to $\mathbf{W}_q$ and $\mathbf{W}_v$, we refer to this variant as the QV-Expert configuration.}

The modified forward computations for these components are given by:
\begin{equation}
    \mathbf{h}^\prime = \mathbf{e} + \text{MLP}{\left( \mathbf{e} \right)} +  \mathbf{E}_{t}^{MLP}\left({\mathbf{e}}\right),\\
    \label{eq:mlp_expert}
\end{equation}
\begin{equation}
    \mathbf{h}^\prime =\text{Attn}\left(\mathbf{h}_Q+\mathbf{E}_{t}^{Q}\left({\mathbf{e}}\right),\mathbf{h}_K+\mathbf{E}_{t}^{K}\left({\mathbf{e}}\right),\mathbf{h}_V+\mathbf{E}_{t}^{V}\left({\mathbf{e}}\right)\right),
    \label{eq:qkv_expert}
\end{equation}
where $\mathbf{e}$ and $\mathbf{h}$ are the inputs and outputs of the original module, respectively. \textcolor{red}{Here, $\mathbf{E}_t$ denotes the task-\(t\) LoRA-Expert module rather than a complete independent ViT. It is a group of LoRA adapters inserted into the selected branch or projection, and each historical expert \(\mathbf{E}_s\), \(s<t\), is frozen after its task has been learned.} The attention operation is defined as:
\begin{equation}
    \text{Attn}\left(\mathbf{Q}, \mathbf{K}, \mathbf{V}_{attn} \right) = \text{softmax}\left(\frac{\mathbf{Q} \mathbf{K}^\top}{\sqrt{d}}\right) \mathbf{V}_{attn},\\
    \label{eq:attn}
\end{equation}
\textcolor{red}{Here, \(\mathbf{V}_{attn}\) denotes the attention value matrix and is distinct from the LoRA matrix \(\mathbf{V}_t\). For \(N\) tokens, the softmax term is an attention-weight matrix \(\mathbf{A}\in\mathbf{R}^{N\times N}\), so \(\mathbf{A}\mathbf{V}_{attn}\) is the standard weighted sum over value vectors.}
The multi-head extension is omitted for clarity. \textcolor{red}{Each LoRA-Expert shares the same architecture but learns distinct parameters. Under the row-vector convention, for an input \(\mathbf{e}\in\mathbf{R}^{1\times d_{in}}\), one adapter consists of low-rank matrices \(\mathbf{U}_t \in \mathbf{R}^{d_{in} \times r}\) and \(\mathbf{V}_t \in \mathbf{R}^{r \times d_{out}}\), producing:}
\begin{equation}
    \mathbf{E}_{t}\left({\mathbf{e}}\right) = \mathbf{e}\mathbf{U}_{t}\mathbf{V}_{t}.
    \label{eq:expert_output}
\end{equation}
\textcolor{red}{In the default ViT-B/16 QV-Expert setting, \(d_{in}=d_{out}=768\) and \(r=10\); for the MLP-Expert variant, \(d_{in}\) and \(d_{out}\) correspond to the inserted MLP branch dimensions.}

In the ViT setting, we denote by $\phi(\mathbf{x}; \mathbf{E}_{t})$ the output embeddings produced by the PTM equipped with the task-specific expert $\mathbf{E}_{t}$. For the first incremental task, we follow the standard LoRA initialization~\cite{hu2022lora}, setting $\mathbf{V}$ to zero and initializing $\mathbf{U}$ via Kaiming initialization~\cite{he2015delving}. For subsequent tasks, we initialize each new LoRA-Expert by copying the weights from the preceding expert, then fine-tune it while keeping all previous experts frozen. As illustrated in Figure~\ref{DLEPEM}(a), this strategy ensures that each task-specific expert adapts independently, preserving knowledge from prior stages. 
\textcolor{red}{This design provides a direct stability-plasticity mechanism. Let $\theta_t=\{\mathbf{U}_t,\mathbf{V}_t\}$ denote the LoRA parameters of the expert allocated to task $t$, while the PTM weights $\mathbf{W}$ remain frozen. After task $s$ has been learned, its old expert parameters $\theta_s$ are not optimized when learning any later task $t>s$; hence}
\begin{equation}
\textcolor{red}{\nabla_{\theta_s}\mathcal{L}(\mathcal{D}_t)=0,\qquad s<t.}
\label{eq:old_expert_stationary}
\end{equation}
\textcolor{red}{Therefore, old experts remain parameter-stationary during subsequent training, which reduces forgetting caused by repeated modification of shared PEFT parameters. Meanwhile, the current expert $\theta_t$ is optimized with the cross-entropy objective in Eq.~(\ref{ce_loss}) without imposing gradient-orthogonality or update-magnitude shrinking constraints, preserving plasticity for newly introduced classes. The remaining source of old-task degradation is mainly expert retrieval error, which is addressed by the prototype-ensemble matching mechanism.}

\subsection{Prototype-Ensemble Matching Mechanism}
When the distribution gap between the pre-trained model (PTM) and the downstream dataset is small, generalized features extracted by the frozen PTM can enable effective module-sample matching. However, since the relationship between pre-training and downstream distributions is generally unknown, it is essential to adaptively capture domain-specific features from the downstream task.

To tackle this, we introduce a dynamically updated router $\mathbf{E}^{router}$ that extracts domain-specific features. The router shares the same architecture as the LoRA-Experts but is continuously updated across incremental tasks.

Under typical incremental constraints, where only data from the current task is available, directly fine-tuning the router leads to overfitting, causing it to route all samples to the most recent LoRA-Expert. To mitigate this, we propose dual feature distillation mechanisms that jointly enforce stability (retaining prior knowledge) and plasticity (adapting to new tasks), regularizing router updates to ensure robust generalization.

\textcolor{red}{\textbf{1. Plasticity Feature Distillation.}} \textcolor{red}{To encourage the router to learn category-specific features of the current task, we first compute the LoRA-Expert prototype for each class \(i\in\mathcal{Y}_t\):}
\begin{equation} \label{eq:prototype_computation}
\textcolor{red}{
\mathbf{P}_{i,t}^{L}=\frac{1}{N_{i,t}}\sum_{(\mathbf{x}_j,\mathbf{y}_j)\in\mathcal{D}_t}
\mathbb{I}(\mathbf{y}_j=i)\phi(\mathbf{x}_j;\mathbf{E}_{t}),\quad
N_{i,t}=\sum_{(\mathbf{x}_j,\mathbf{y}_j)\in\mathcal{D}_t}\mathbb{I}(\mathbf{y}_j=i).
}
\end{equation}
\textcolor{red}{Here, \(N_{i,t}\) is the number of current-task samples belonging to class \(i\), and \(\mathbb{I}(\cdot)\) denotes the indicator function. We then align the router feature of each current-task sample with the LoRA-Expert prototype of its ground-truth class using a KL-divergence objective:}
\begin{equation}  \label{eq:pla_fd}
\textcolor{red}{
\mathcal{L}_{PFD}=\frac{1}{|\mathcal{D}_t|}\sum_{(\mathbf{x}_j,\mathbf{y}_j)\in\mathcal{D}_t}
\mathrm{KL}\left(
\sigma(\mathbf{P}_{\mathbf{y}_j,t}^{L}/\tau)\,\|\,\sigma(\phi(\mathbf{x}_j;\mathbf{E}^{router}_{t})/\tau)
\right).
}
\end{equation}
\textcolor{red}{Here, \(\sigma(\mathbf{z})=\mathrm{softmax}(\mathbf{z})\), and temperature scaling is written explicitly as \(\sigma(\mathbf{z}/\tau)=\mathrm{softmax}(\mathbf{z}/\tau)\), where \(\tau\) is the temperature parameter. Thus, \(\sigma(\cdot)\) in the distillation losses denotes the same softmax function as \(\mathrm{softmax}(\cdot)\) in the attention equation. These class-specific prototypes guide the router toward discriminative features of the current task, ensuring effective adaptation for new tasks.}


\textcolor{red}{\textbf{2. Stability Feature Distillation.}} 
To mitigate the degradation of previously learned knowledge, \textcolor{red}{we introduce a stability constraint that encourages the current router to mimic the feature distribution produced by the previous router on the current-task inputs:}
\begin{equation}  \label{eq:sta_fd}
\textcolor{red}{
\mathcal{L}_{SFD}=\frac{1}{|\mathcal{D}_t|}\sum_{(\mathbf{x}_j,\mathbf{y}_j)\in\mathcal{D}_t}
\mathrm{KL}\left(
\sigma(\phi(\mathbf{x}_j;\mathbf{E}^{router}_{t-1})/\tau)\,\|\,\sigma(\phi(\mathbf{x}_j;\mathbf{E}^{router}_{t})/\tau)
\right).
}
\end{equation}

The overall router loss combines these objectives: 
\begin{equation}
\textcolor{red}{
\mathcal{L}_{router} =
\begin{cases}
\mathcal{L}_{PFD}, & t=1,\\
\alpha\mathcal{L}_{PFD}+(1-\alpha)\mathcal{L}_{SFD}, & t>1,
\end{cases}
}
    \label{router_loss}
\end{equation}
\textcolor{red}{where \(\alpha\) weights the plasticity and stability distillation terms for the router. For the first task, \(\mathcal{L}_{SFD}\) is omitted because no previously trained router exists. The router parameters \(\mathbf{E}^{router}_{t}\) are optimized by minimizing this combined loss function. With the default \(\alpha=0.04\), the router update places more weight on \(\mathcal{L}_{SFD}\); however, this coefficient controls router regularization only and should not be interpreted as proof of an optimal global stability-plasticity balance. The main source of new-task plasticity remains the trainable current LoRA-Expert, while frozen old experts provide parameter-level stability.}

At the end of each incremental stage, we compute general prototypes $\mathbf{P}^{F}_{i}$ using the frozen PTM $\phi(\mathbf{x})$ and domain-specific prototypes $\mathbf{P}^{R}_{i}$ using the current router $\mathbf{E}^{router}_{t}$ for each newly introduced class $i\in\mathcal{Y}_t$, as defined in Eq.~(\ref{eq:prototype_computation}). \textcolor{red}{The prototype dictionary is updated in an append-only manner: old entries for classes \(i\in\mathbf{Y}_{t-1}\) are retained as historical keys and are not recomputed with later-task data. The new ensemble keys created at stage \(t\) are}
\begin{equation}  \label{eq:prototype_ensemble}
\textcolor{red}{
	\mathbf{K}^{new}_{t}= \left\{\mathbf{K}_i=[\mathbf{P}^{F}_{i};\mathbf{P}^{R}_{i}] \mid i\in\mathcal{Y}_t\right\}.}
\end{equation}
\textcolor{red}{Equivalently, we define the class-\(i\) ensemble prototype vector as \(\mathbf{P}_i=\mathrm{concat}(\mathbf{P}^{F}_{i},\mathbf{P}^{R}_{i})=[\mathbf{P}^{F}_{i};\mathbf{P}^{R}_{i}]\), and use it as the dictionary key \(\mathbf{K}_i=\mathbf{P}_i\). The semicolon in \([\cdot;\cdot]\) denotes vector concatenation, not matrix addition or summation.}
\textcolor{red}{Each new key \(\mathbf{K}_i\) is appended to \(\text{Dict}\) together with the expert index \(\nu_i=t\), which points to the corresponding expert \(\mathbf{E}_t\). Thus, after task \(t\), the dictionary covers all seen classes \(\mathbf{Y}_t\), while only entries for \(\mathcal{Y}_t\) are newly created. This keeps DLEPEM rehearsal-free: the persistent state consists of frozen LoRA-Experts, the current router, and compact ensemble keys, but no raw samples, old mini-batches, feature buffers, or exemplar sets.}

\textcolor{red}{Because the router is updated across tasks, router-domain keys from older stages may have been generated by earlier router states. DLEPEM mitigates this router-state drift in two ways. First, each key includes the frozen-PTM component $\mathbf{P}^{F}_{i}$, which remains comparable across stages because the PTM is fixed. Second, stability feature distillation in Eq.~(\ref{eq:sta_fd}) encourages $\mathbf{E}^{router}_{t}$ to preserve the behavior of $\mathbf{E}^{router}_{t-1}$ while adapting to the current task.}

\paragraph{Prototype-Ensemble Matching} Our mechanism dynamically selects the most suitable LoRA-Expert for each input by leveraging a key-value association strategy inspired by L2P~\cite{wang2022learning}. \textcolor{red}{Specifically, we maintain a dictionary where each seen class \(i\in\mathbf{Y}_t\) has one ensemble key \(\mathbf{K}_i\) and an associated expert index \(\nu_i\):}
\begin{equation}  \label{eq:key}
\textcolor{red}{
\text{Dict}_{t}=\left\{\left(\mathbf{K}_i,\nu_i\right)\mid i\in\mathbf{Y}_t,\ \nu_i\in\{1,\ldots,t\}\right\}.}
\end{equation}
\textcolor{red}{Here, \(\nu_i\) identifies the task-specific LoRA-Expert associated with class \(i\).}

\textcolor{red}{During inference, given an input $\mathbf{x}$, we construct an ensemble query $\mathbf{q}(\mathbf{x}) = \mathrm{concat}(\mathbf{P}^F, \mathbf{P}^R)=[\mathbf{P}^F;\mathbf{P}^R]$, where $\mathbf{P}^F$ is the feature from the frozen PTM and $\mathbf{P}^R$ is the domain-specific feature from the router. We then identify the nearest key by cosine similarity:}
\begin{equation}
    \hat{i} = \underset{i \in \mathbf{Y}_t}{\text{argmax}} \left( \text{cos}(\mathbf{q}(\mathbf{x}), \mathbf{K}_i) \right),
    \label{prototype_matching}
\end{equation}
\textcolor{red}{The retrieved key \(\mathbf{K}_{\hat{i}}\) returns an expert identity rather than a direct class prediction. Let \(\hat{t}=\nu_{\hat{i}}\) denote the task index associated with \(\mathbf{K}_{\hat{i}}\) in \(\text{Dict}_{t}\); DLEPEM then selects \(\mathbf{E}_{\hat{t}}\) for within-task prediction. Final classification is restricted to the class set \(\mathcal{Y}_{\hat{t}}\) associated with the selected expert and uses the corresponding prototype weights. Thus, all stored ensemble keys participate in module-identity inference, while prototype weights from unrelated experts are not mixed during classification. Algorithm~\ref{alg:dlepem_inference} summarizes this inference pipeline.}

\subsection{Optimization Objective and Training Procedure}
DLEPEM employs a two-stage training paradigm: it first learns Dynamic LoRA-Expert modules and then optimizes the router for effective module-sample matching. Algorithm~\ref{alg:dlepem} provides the complete training procedure, and Algorithm~\ref{alg:dlepem_inference} gives the corresponding task-agnostic inference procedure.

\textbf{1. Dynamic LoRA-Expert Learning:}
The objective function for training the LoRA-Expert is:
\begin{equation}
   \min_{\mathbf{W}_{cls}, \mathbf{E}_i}\text{L}_{CE}\left(\mathbf{W}_{cls}^\top\phi\left(\mathbf{x};\mathbf{E}_i\right),\mathbf{y}\right)
    \label{ce_loss},
\end{equation}
where $\text{L}_{CE}$ denotes the cross-entropy loss, and $\phi(\mathbf{x}; \mathbf{E}_i)$ represents the PTM equipped with the LoRA-Expert $\mathbf{E}_i$. \textcolor{red}{During inference, DLEPEM uses LoRA-Expert-derived prototype weights for classification. Specifically, $\mathbf{P}^L$ denotes the prototype-weight matrix formed by the class prototypes $\mathbf{P}_{i,t}^{L}$ defined in Eq.~(\ref{eq:prototype_computation}).} Classification is performed using cosine similarity:
\begin{equation} \label{eq:final_fx}
\text{f}(\mathbf{x}|\mathbf{E}_i)=(\frac {\mathbf{P}^L}{\|\mathbf{P}^L\|_2})^\top(\frac{\phi(\mathbf{x};\mathbf{E}_i)}{\|\phi(\mathbf{x};\mathbf{E}_i)\|_2}),
\end{equation}
where $\mathbf{E}_i$ denotes the selected LoRA-Expert and $\mathbf{P}^L$ contains the prototype weights associated with this expert.

\begin{algorithm}[H]
    \caption{Training Procedure for DLEPEM}
    \label{alg:dlepem}
    \begin{algorithmic}[1]
    \STATE \textbf{Input:} Pre-trained model $\phi(\cdot)$, incremental datasets $\{\mathcal{D}_1, \ldots, \mathcal{D}_T\}$, LoRA rank $r$, distillation coefficient $\alpha$.
    \STATE \textbf{Output:} Trained LoRA-Experts $\{\mathbf{E}_1, \ldots, \mathbf{E}_T\}$, Router $\mathbf{E}^{router}_T$, Prototype Dictionary $\text{Dict}$.
    \STATE \textbf{Initialize:} $\text{Dict} \leftarrow \emptyset$, $\mathbf{E}^{router}_0$ with random weights.
    \FOR{$t = 1$ \textbf{to} $T$}
        \STATE \textcolor{blue}{\# Stage 1: Dynamic LoRA-Expert Learning}
        \STATE Initialize LoRA-Expert $\mathbf{E}_t$. If $t>1$, copy weights from $\mathbf{E}_{t-1}$.
        \STATE Train $\mathbf{E}_t$ and classifier $\mathbf{W}_{cls}$ on $\mathcal{D}_t$ using $\mathcal{L}_{CE}$ (Eq. (\ref{ce_loss})).
        \STATE Freeze parameters of $\mathbf{E}_t$.
        
        \STATE \textcolor{blue}{\# Stage 2: Router Learning and Prototype-Ensemble Building}
        \STATE \textcolor{red}{If \(t>1\), freeze a copy of the previous router \(\mathbf{E}^{router}_{t-1}\).}
        \STATE Train router $\mathbf{E}^{router}_t$ on $\mathcal{D}_t$ using $\mathcal{L}_{router}$ (Eq. (\ref{router_loss})).
        
        \STATE \textcolor{blue}{\# Append Prototype Dictionary with newly introduced classes}
        \STATE \textcolor{red}{Keep old dictionary entries for classes in $\mathbf{Y}_{t-1}$ unchanged.}
        \FOR{each newly introduced class $i \in \mathcal{Y}_t$}
            \STATE Compute general prototype $\mathbf{P}^{F}_{i}$ using frozen PTM $\phi(\cdot)$ and samples from $\mathcal{D}_t$.
            \STATE Compute domain-specific prototype $\mathbf{P}^{R}_{i}$ using router $\mathbf{E}^{router}_{t}$ and samples from $\mathcal{D}_t$.
            \STATE \textcolor{red}{Create ensemble prototype $\mathbf{P}_i = [\mathbf{P}^{F}_{i}; \mathbf{P}^{R}_{i}]$ (Eq. (\ref{eq:prototype_ensemble})).}
            \STATE \textcolor{red}{Append an entry to $\text{Dict}$ by associating key $\mathbf{K}_i = \mathbf{P}_i$ with expert index \(\nu_i=t\) (Eq. (\ref{eq:key})).}
        \ENDFOR
    \ENDFOR
    \STATE \textbf{return} $\{\mathbf{E}_1, \ldots, \mathbf{E}_T\}$, $\mathbf{E}^{router}_T$, $\text{Dict}$.
    \end{algorithmic}
\end{algorithm}

\begin{algorithm}[H]
    \caption{\textcolor{red}{Inference Procedure for DLEPEM}}
    \label{alg:dlepem_inference}
    \begin{algorithmic}[1]
    \STATE \textcolor{red}{\textbf{Input:} Test sample $\mathbf{x}$, frozen PTM $\phi(\cdot)$, frozen LoRA-Experts $\{\mathbf{E}_1,\ldots,\mathbf{E}_T\}$, current router $\mathbf{E}^{router}_T$, prototype dictionary $\text{Dict}$, and prototype weights $\{\mathbf{P}^{L}_{c}\}_{c\in\mathbf{Y}_T}$.}
    \STATE \textcolor{red}{\textbf{Output:} Predicted label $\hat{y}$.}
    \STATE \textcolor{red}{Compute frozen-PTM feature $\mathbf{P}^{F}(\mathbf{x})=\phi(\mathbf{x})$.}
    \STATE \textcolor{red}{Compute router-domain feature $\mathbf{P}^{R}(\mathbf{x})=\phi(\mathbf{x};\mathbf{E}^{router}_T)$.}
    \STATE \textcolor{red}{Build the ensemble query $\mathbf{q}(\mathbf{x})=[\mathbf{P}^{F}(\mathbf{x}),\mathbf{P}^{R}(\mathbf{x})]$.}
    \STATE \textcolor{red}{Retrieve the nearest dictionary key $\hat{i}=\arg\max_{i\in\mathbf{Y}_T}\text{cos}(\mathbf{q}(\mathbf{x}),\mathbf{K}_i)$.}
    \STATE \textcolor{red}{Obtain the expert index \(\hat{t}=\nu_{\hat{i}}\) associated with $\mathbf{K}_{\hat{i}}$ in $\text{Dict}$.}
    \STATE \textcolor{red}{Extract the selected-expert feature $\mathbf{z}=\phi(\mathbf{x};\mathbf{E}_{\hat{t}})$.}
    \STATE \textcolor{red}{Restrict candidate labels to $\mathcal{Y}_{\hat{t}}$ and predict $\hat{y}=\arg\max_{c\in\mathcal{Y}_{\hat{t}}}\text{cos}(\mathbf{z},\mathbf{P}^{L}_{c})$.}
    \STATE \textcolor{red}{\textbf{return} $\hat{y}$.}
    \end{algorithmic}
\end{algorithm}

\textbf{2. Router Learning:}
The router is optimized with the objective function shown in Eq.~(\ref{router_loss}).

As illustrated in Figure~\ref{DLEPEM}, we divide the class-incremental learning process into two stages:
\textcolor{red}{First, a new LoRA-Expert is learned for each incremental task to capture task-specific features; each expert is categorized as either an MLP-Expert or a QV-Expert depending on its insertion point. Second, we introduce a prototype-ensemble matching mechanism that captures both general and domain-specific features, thereby improving module-sample matching. Notably, DLEPEM's components are orthogonal to many existing approaches and can be integrated with them straightforwardly.}

\section{Experiments}
\subsection{Experimental Settings}

\textbf{Datasets:}
\textcolor{red}{We evaluate DLEPEM in two settings: CIL and Few-Shot Class-Incremental Learning (FSCIL)~\cite{tao2020few}.} For CIL, we follow standard protocols~\cite{zhou2023revisitingclassincrementallearningpretrained,wu2025sdlora} and test on five benchmarks: VTAB~\cite{zhai2019large}, CIFAR100~\cite{krizhevsky2009learning}, CUB200~\cite{wah2011caltech}, ImageNet-R~\cite{hendrycks2021many}, and OmniBenchmark~\cite{zhang2022benchmarking}. \textcolor{red}{VTAB comprises 50 classes, CIFAR100 contains 100 classes, CUB200 and ImageNet-R each contain 200 classes, and OmniBenchmark, the largest benchmark among them, includes 300 classes.} As shown in Table~\ref{tab:cil-config}, we follow the common practices~\cite{gao2023unified, liang2024inflora, wu2025sdlora}, splitting CIFAR100 into 10 tasks, CUB200 into 10 tasks, ImageNet-R into 5 tasks, OmniBenchmark into 10 tasks, and VTAB into 5 tasks.

\begin{table}[t]
    \centering
    \caption{\textcolor{red}{Configuration of the class-incremental learning (CIL) benchmarks.}}
    \label{tab:cil-config}
    \resizebox{\linewidth}{!}{
        \begin{tabular}{lccccc}
            \toprule
            Task                & CIFAR100 & CUB200 & ImageNet-R & OmniBenchmark & VTAB \\
            \midrule
            Classes\ /\ Task    & 10       & 20     & 40         & 30            & 10   \\
            \# of tasks         & 10       & 10     & 5          & 10            & 5    \\
            \bottomrule
        \end{tabular}
    }
\end{table}
\begin{table}[t]
    \centering
    \caption{\textcolor{red}{Configuration of the few-shot class-incremental learning (FSCIL) benchmarks.}}
    \label{tab:fscil-config}
    \resizebox{\linewidth}{!}{
        \begin{tabular}{lccc}
            \toprule
            Task                & CUB200        & CIFAR100      & \textit{mini}ImageNet \\
            \midrule
            Base Classes        & 100           & 60            & 60                    \\
            Incremental Tasks   & 10-way 5-shot & 5-way 5-shot  & 5-way 5-shot          \\
            \# of Tasks         & 1+10          & 1+8           & 1+8                   \\
            \bottomrule
        \end{tabular}
    }
\end{table}

\textcolor{red}{For FSCIL, we adopt the settings used in prior work~\cite{park2024pre,liu2024few} on CUB200, CIFAR100, and \textit{mini}ImageNet~\cite{russakovsky2015imagenet}.} As shown in Table~\ref{tab:fscil-config}, we use 100 classes in CUB200 as the base class set for the first task. The remaining 100 classes are partitioned into 10 incremental tasks, with each incremental task containing 10 new classes and the few-shot training set containing 5 examples per class (10-way 5-shot incremental task). CIFAR100 and \textit{mini}ImageNet are divided into 60 classes for the base task, and the remaining 40 classes are divided into eight 5-way 5-shot incremental tasks.

\textbf{Comparison methods:} 
\textcolor{red}{For CIL, we compare DLEPEM against several representative and recent methods, including SimpleCIL~\cite{zhou2023revisitingclassincrementallearningpretrained}, prompt-based methods (L2P~\cite{wang2022learning}, DualPrompt~\cite{wang2022dualprompt}, CODA-Prompt~\cite{smith2023coda}), LoRA-based methods (LAE~\cite{gao2023unified}, APER~\cite{zhou2023revisitingclassincrementallearningpretrained}, InfLoRA~\cite{liang2024inflora}, SD-LoRA~\cite{wu2025sdlora}, BiLoRA~\cite{zhu2025bilora}), and recent PTM-based CIL methods such as EASE~\cite{zhou2024ease}.} We also include standard full fine-tuning as a baseline, where the model is sequentially fine-tuned without any continual learning mechanism. \textcolor{red}{We reproduce all baseline results in our benchmark tables under the corresponding CIL/FSCIL protocols. For fairness, all methods use the same pre-trained backbone and identical data splits; for LoRA/adapter-based methods, we set the rank to \(r=10\) where applicable. Other method-specific training settings, including optimizer, training epochs, batch size, and augmentation policy, follow their original or recommended implementations.}

For FSCIL, we additionally benchmark against three recent ViT-based methods tailored for few-shot scenarios: PriViLege~\cite{park2024pre}, ASP~\cite{liu2024few}, and CPE-CLIP~\cite{10350931}. All methods leverage the same pre-trained backbone (ViT-B/16-IN21K~\cite{dosovitskiy2020image}) and identical data splits to ensure fair comparisons. 

\textbf{Evaluation metrics:} For CIL, we assess model performance using two established metrics: $\bar{\mathcal{A}} = \frac{1}{T}\sum_{i=1}^{T}\mathcal{ACC}_{i}$ and $\mathcal{A}_L$~\cite{liang2024inflora}. Here, $\bar{\mathcal{A}}$ is the average accuracy of all $T$ incremental stages, and $\mathcal{A}_L$ is the accuracy of the last incremental stage. $\mathcal{ACC}_{i}$ is defined as:
\begin{equation}
    {\mathcal{ACC}_{i}} = \frac{1}{i} \sum_{j=1}^{i}a_{i,j},
    \label{ave_acc}
\end{equation}
where $a_{i,j}$ denotes the accuracy on the  
$j$-th task after training on the $i$-th task. For FSCIL, $\mathcal{A}_{\text{Base}}$ is the accuracy of the base classes in task 0. Both $\mathcal{A}_L$ and $\bar{\mathcal{A}}$ are defined identically to those in the standard CIL setting.

\begin{table}[t]\scriptsize
    \centering
    \caption{\textcolor{red}{Performance comparison on CIL benchmarks using the same ViT-B/16-IN21K backbone. Results are reported as mean \(\pm\) standard deviation over three runs; the best result in each column is highlighted in bold.}}
    \label{tab:cil_benchmarks}
    \resizebox{1.0\linewidth}{!}{
    \begin{tabular}{@{}l *{5}{cc} c@{}}
        \toprule
        \multirow{2}{*}{Method} & 
        \multicolumn{2}{c}{CIFAR100 ($T$=10)} & 
        \multicolumn{2}{c}{CUB200 ($T$=10)} & 
        \multicolumn{2}{c}{ImageNet-R ($T$=5)} & 
        \multicolumn{2}{c}{OmniBenchmark ($T$=10)} & 
        \multicolumn{2}{c}{VTAB ($T$=5)} \\
        \cmidrule(lr){2-3} \cmidrule(lr){4-5} \cmidrule(lr){6-7} \cmidrule(lr){8-9} \cmidrule(lr){10-11}
        & $\mathcal{A}_L$ & $\bar{\mathcal{A}}$ 
        & $\mathcal{A}_L$ & $\bar{\mathcal{A}}$ 
        & $\mathcal{A}_L$ & $\bar{\mathcal{A}}$ 
        & $\mathcal{A}_L$ & $\bar{\mathcal{A}}$ 
        & $\mathcal{A}_L$ & $\bar{\mathcal{A}}$ \\
        \midrule
        Full Fine-Tuning 
            & $66.26 \pm 0.18$ & $76.94 \pm 0.02$ 
            & $55.29 \pm 0.46$ & $70.30 \pm 0.74$  
            & $59.90 \pm 0.05$ & $72.19 \pm 0.12$ 
            & $47.75 \pm 0.14$ & $65.86 \pm 0.12$ 
            & $62.95 \pm 5.94$ & $80.80 \pm 1.50$ \\
        
        SimpleCIL~\cite{zhou2023revisitingclassincrementallearningpretrained} 
            & $81.27 \pm 0.01$ & $87.13 \pm 0.01$ 
            & $82.28 \pm 6.35$ & $91.85 \pm 0.00$ 
            & $65.14 \pm 15.29$ & $59.72 \pm 0.03$ 
            & $66.98 \pm 8.93$ & $79.35 \pm 0.01$
            & $80.74 \pm 5.26$ & $90.80 \pm 0.00$ \\
        
        L2P~\cite{wang2022learning}
            & $84.82 \pm 0.22$ & $89.78 \pm 0.01$ 
            & $71.98 \pm 0.07$ & $81.80 \pm 0.03$ 
            & $72.08 \pm 0.01$ & $76.76 \pm 0.06$ 
            & $64.45 \pm 0.07$ & $74.14 \pm 0.03$ 
            & $64.27 \pm 0.02$ & $81.84 \pm 0.19$ \\
        
        DualPrompt~\cite{wang2022dualprompt} 
            & $85.23 \pm 0.07$ & $90.32 \pm 0.06$ 
            & $74.23 \pm 0.10$ & $84.81 \pm 0.02$ 
            & $69.34 \pm 0.09$ & $73.58 \pm 0.05$ 
            & $66.16 \pm 0.06$ & $74.97 \pm 0.03$ 
            & $78.90 \pm 0.10$ & $89.82 \pm 0.06$ \\
        
        CODA-Prompt~\cite{smith2023coda} 
            & $86.69 \pm 0.00$ & $91.31 \pm 0.01$ 
            & $75.45 \pm 0.00$ & $84.65 \pm 0.00$ 
            & $75.16 \pm 0.08$ & $80.46 \pm 0.04$ 
            & $68.67 \pm 0.00$ & $77.79 \pm 0.00$ 
            & $75.08 \pm 0.00$ & $87.24 \pm 0.00$ \\
            
        APER~\cite{zhou2023revisitingclassincrementallearningpretrained} 
            &$87.32 \pm 0.01$ &$92.09 \pm 0.02$ 
            &$86.82 \pm 0.04$ &$91.84 \pm 0.03$ 
            &$68.22 \pm 0.07$ &$75.30 \pm 0.07$ 
            &$74.40 \pm 0.01$ &$80.62 \pm 0.01$ 
            &$84.44 \pm 0.01$ &$86.27 \pm 0.02$ \\
        
        LAE~\cite{gao2023unified} 
            &$85.60 \pm 0.19$ &$91.26 \pm 0.12$ 
            &$67.91 \pm 0.04$ &$80.17 \pm 0.04$ 
            &$71.26 \pm 0.18$ &$77.08 \pm 0.04$ 
            &$66.14 \pm 0.06$ &$74.89 \pm 0.03$ 
            &$68.45 \pm 2.53$ &$85.31 \pm 0.44$ \\
        
        InfLoRA~\cite{liang2024inflora} 
            &$86.43 \pm 0.02$ &$91.80 \pm 0.02$ 
            &$70.07 \pm 0.26$ &$81.71 \pm 0.17$ 
            &$77.66 \pm 0.08$ &$82.90 \pm 0.05$ 
            &$68.38 \pm 0.12$ &$78.06 \pm 0.03$ 
            &$74.66 \pm 0.43$ &$86.03 \pm 0.17$ \\
        
        SD-LoRA~\cite{wu2025sdlora} 
            &$87.62 \pm 0.00$ &$92.10 \pm 0.00$ 
            &$72.69 \pm 0.00$ &$83.17 \pm 0.00$ 
            &$78.52 \pm 0.00$ &$82.74 \pm 0.00$
            &$69.32 \pm 0.00$ &$77.78 \pm 0.00$ 
            &$67.73 \pm 0.00$ &$84.44 \pm 0.00$ \\
        \textcolor{red}{EASE~\cite{zhou2024ease}}
            &\textcolor{red}{$88.13 \pm 0.04$} &\textcolor{red}{$92.59 \pm 0.05$}
            &\textcolor{red}{$84.18 \pm 0.06$} &\textcolor{red}{$90.20 \pm 0.04$}
            &\textcolor{red}{$76.95 \pm 0.05$} &\textcolor{red}{$81.48 \pm 0.06$}
            &\textcolor{red}{$67.75 \pm 0.04$} &\textcolor{red}{$74.85 \pm 0.05$}
            &\textcolor{red}{$82.34 \pm 0.06$} &\textcolor{red}{$90.45 \pm 0.04$} \\
        \textcolor{red}{BiLoRA~\cite{zhu2025bilora}}
            &\textcolor{red}{$85.30 \pm 0.05$} &\textcolor{red}{$90.73 \pm 0.04$}
            &\textcolor{red}{$73.75 \pm 0.06$} &\textcolor{red}{$83.67 \pm 0.05$}
            &\textcolor{red}{$76.33 \pm 0.04$} &\textcolor{red}{$81.21 \pm 0.07$}
            &\textcolor{red}{$68.87 \pm 0.05$} &\textcolor{red}{$77.53 \pm 0.04$}
            &\textcolor{red}{$76.76 \pm 0.06$} &\textcolor{red}{$88.94 \pm 0.05$} \\
        \midrule
        DLEPEM-MLP 
            &$85.99 \pm 0.34$ &$92.40 \pm 0.03$ 
            &$\bf{87.70 \pm 0.05}$ &$\bf{92.31 \pm 0.29}$ 
            &$76.25 \pm 0.44$ &$82.37 \pm 0.14$ 
            &$74.32 \pm 0.00$ &$81.60 \pm 0.00$ 
            &$84.96 \pm 0.22$ &$\bf{91.84 \pm 0.03}$ \\
        DLEPEM-QV 
            &$\bf{88.84 \pm 0.09}$ &$\bf{93.39 \pm 0.10}$ 
            &$87.56 \pm 0.16$ &$92.09 \pm 0.08$ 
            &$\bf{78.77 \pm 0.07}$ &$\bf{83.43 \pm 0.12}$ 
            &$\bf{75.53 \pm 0.08}$ &$\bf{82.16 \pm 0.06}$  
            &$\bf{85.18 \pm 0.14}$ &$91.11 \pm 0.13$ \\
        \bottomrule
    \end{tabular}
    }
\end{table}

\textbf{Architecture and training details:} 
We adopt ViT-B/16-IN21K~\cite{dosovitskiy2020image} pre-trained on ImageNet-21K as the backbone. Optimization is conducted using SGD with an initial learning rate of 0.02 and cosine annealing. LoRA-Experts are trained for 20 epochs and the router for 5 epochs, using a batch size of 48. We set the LoRA rank to $r=10$ and insert LoRA-Experts into all Transformer blocks. The distillation coefficient $\alpha$ is set to 0.04. All experiments are performed on an NVIDIA A800 GPU with fixed data splits (seed 1993) and the same backbone to ensure reproducibility; results are averaged over three runs. Following SD-LoRA~\cite{wu2025sdlora}, our QV-Experts are integrated into the query and value projections of the attention module. We also evaluate a variant, DLEPEM-MLP, which introduces MLP-Experts as a parallel branch to the FFN layer.

\subsection{Benchmark Comparison}
\textbf{Class-Incremental Learning:} 
We conduct a comprehensive evaluation of DLEPEM against representative recent methods on five benchmark datasets. As shown in Table~\ref{tab:cil_benchmarks}, DLEPEM consistently delivers superior accuracy across all benchmarks. \textcolor{red}{After adding the recent EASE and BiLoRA baselines, DLEPEM still achieves the best average accuracy on all five CIL benchmarks under our reproduced experimental setting. Compared with the strongest baseline in each dataset, DLEPEM improves $\bar{\mathcal{A}}$ by 0.80\% on CIFAR100, 2.11\% on CUB200, 0.53\% on ImageNet-R, 1.54\% on OmniBenchmark, and 1.39\% on VTAB.}

\begin{figure}[H]
    \centering
    \subfloat[CIFAR100 ($T$=10)\label{fig:benchmark-cil-a}]{%
        \begin{minipage}{0.48\linewidth}
        \centering
        \includegraphics[width=\linewidth]{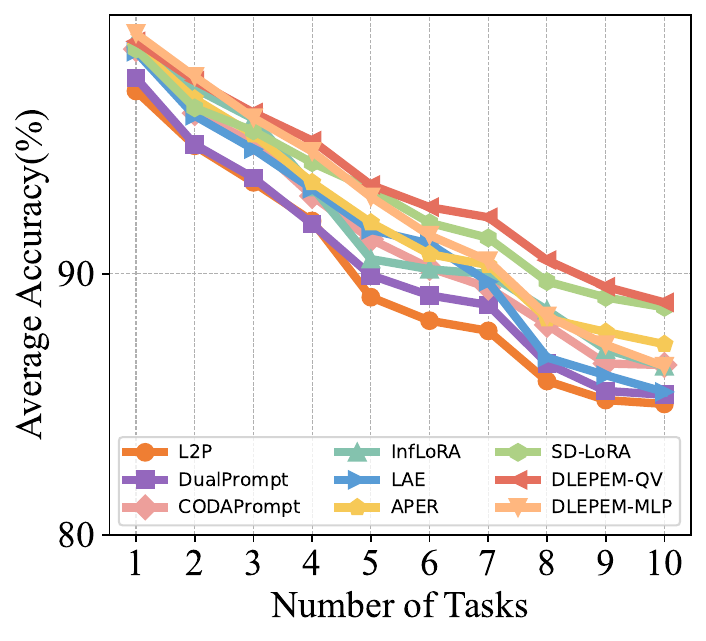}
        \end{minipage}}
    \subfloat[CUB200 ($T$=10)\label{fig:benchmark-cil-b}]{%
        \begin{minipage}{0.48\linewidth}
        \centering
        \includegraphics[width=\linewidth]{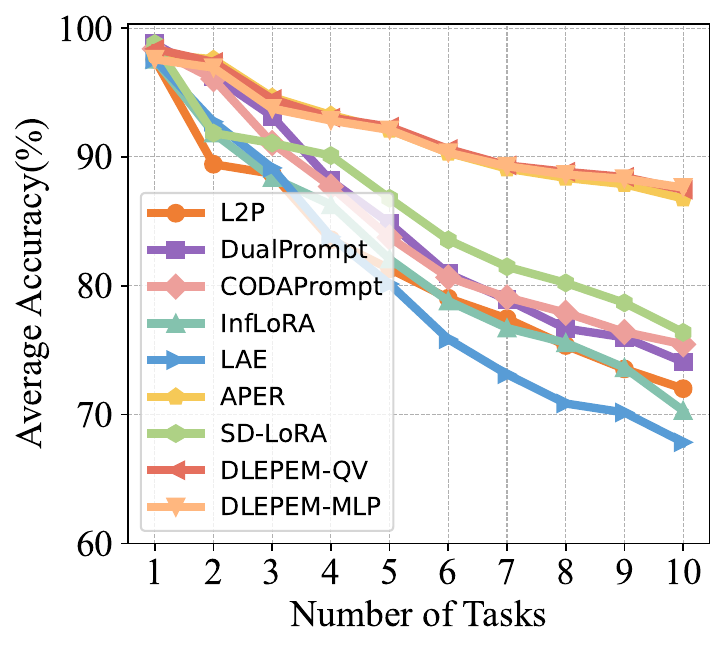}
        \end{minipage}}
    \\
    \subfloat[ImageNet-R ($T$=5)\label{fig:benchmark-cil-c}]{%
        \begin{minipage}{0.48\linewidth}
        \centering
        \includegraphics[width=\linewidth]{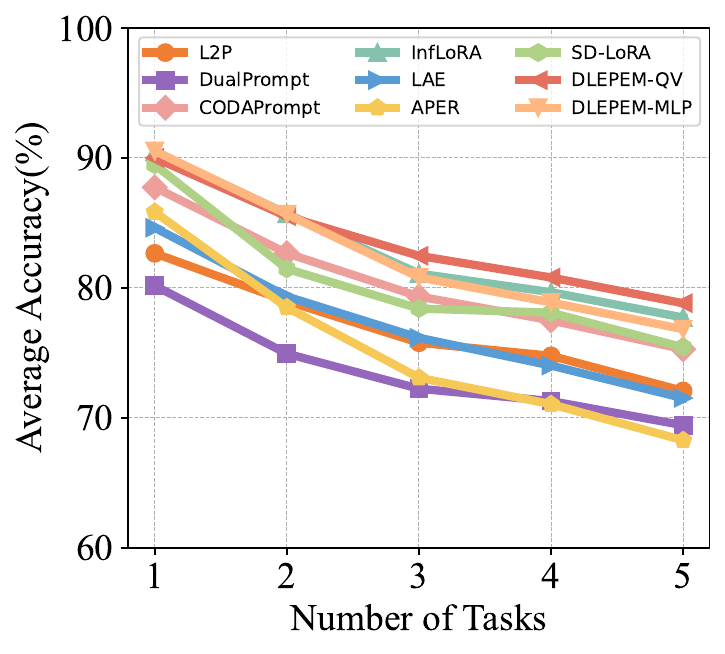}
        \end{minipage}}
    \subfloat[OmniBenchmark ($T$=10)\label{fig:cil_omni10}]{%
        \begin{minipage}{0.48\linewidth}
        \centering
        \includegraphics[width=0.96\linewidth]{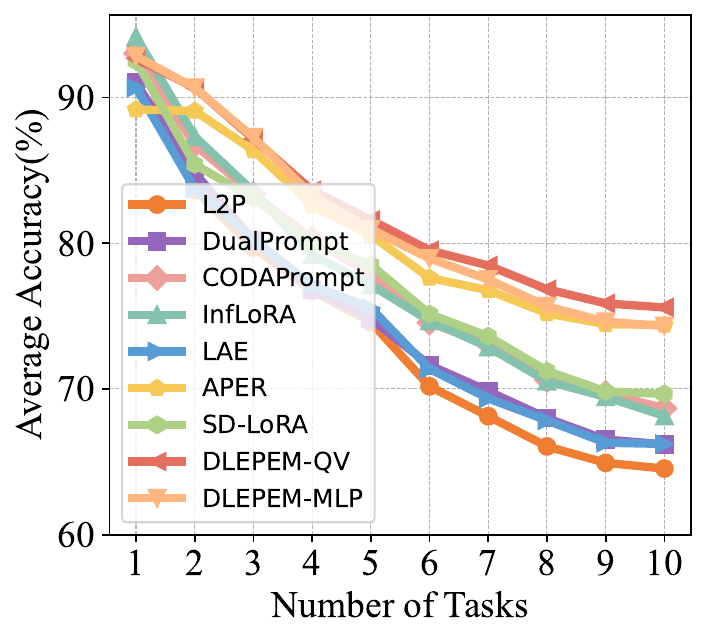}
        \end{minipage}}
    \caption{\textcolor{red}{Incremental accuracy curves on the CIL benchmarks. All methods use the same ViT-B/16-IN21K backbone, and each subplot reports accuracy after successive incremental tasks for the specified dataset protocol.}}
    
    \label{fig:benchmark-cil}
\end{figure}

\textcolor{red}{For instance, on CIFAR100, DLEPEM-QV achieves an average accuracy of 93.39\%, outperforming the newly added EASE baseline by 0.80\%. Following the caution of Kim and Han~\cite{kim2023stabilityplasticity}, we do not interpret high average accuracy alone as proof of an optimal stability-plasticity balance; instead, the old/new-task analysis in Section~\ref{sec:old-new-analysis} provides behavior-level evidence for the trade-off.} Figure~\ref{fig:benchmark-cil} further shows that DLEPEM achieves the highest performance throughout training, underscoring its robustness.

\begin{figure}[t!]
    \centering
    \subfloat[CUB200 ($T$=11)\label{fig:fscil_cub}]{%
        \begin{minipage}{0.32\linewidth}
        \centering
        \includegraphics[width=0.96\linewidth]{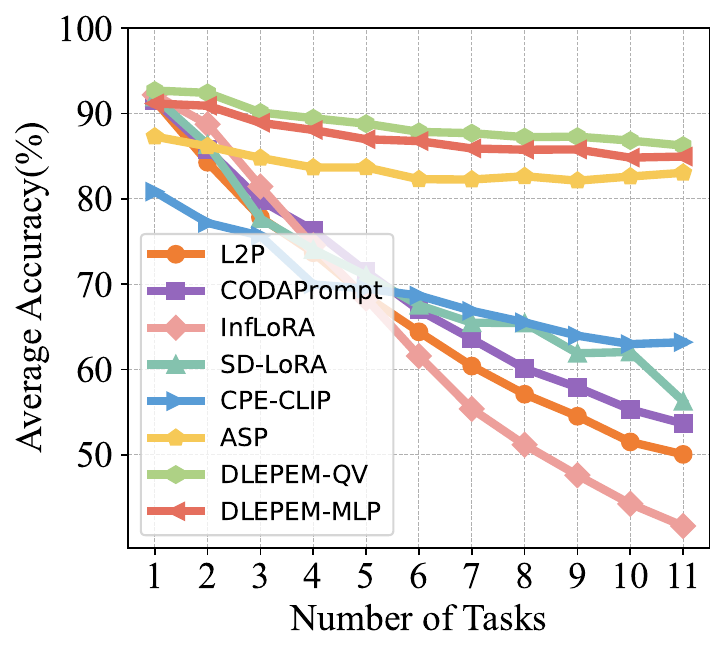}
        \end{minipage}}
    \subfloat[CIFAR100 ($T$=9)\label{fig:fscil_cifar}]{%
        \begin{minipage}{0.32\linewidth}
        \centering
        \includegraphics[width=0.96\linewidth]{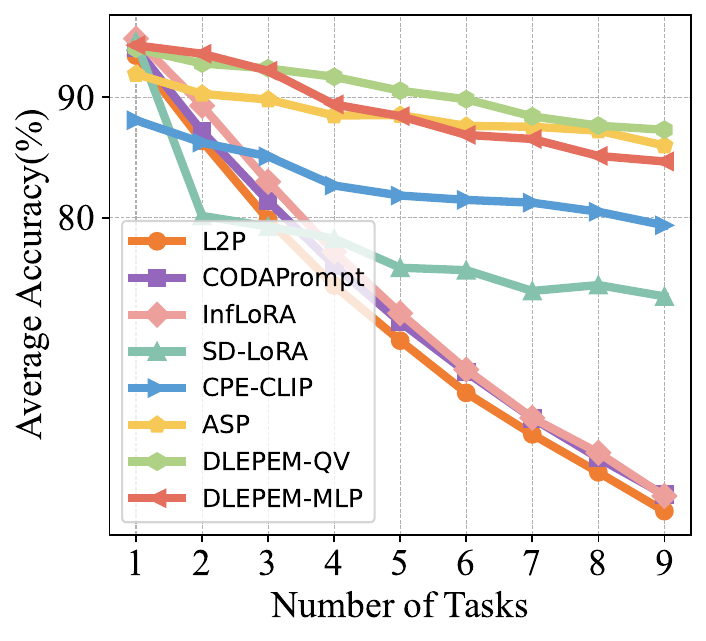}
        \end{minipage}}
    \subfloat[\textit{mini}ImageNet ($T$=9)\label{fig:fscil_mini}]{%
        \begin{minipage}{0.32\linewidth}
        \centering
        \includegraphics[width=0.96\linewidth]{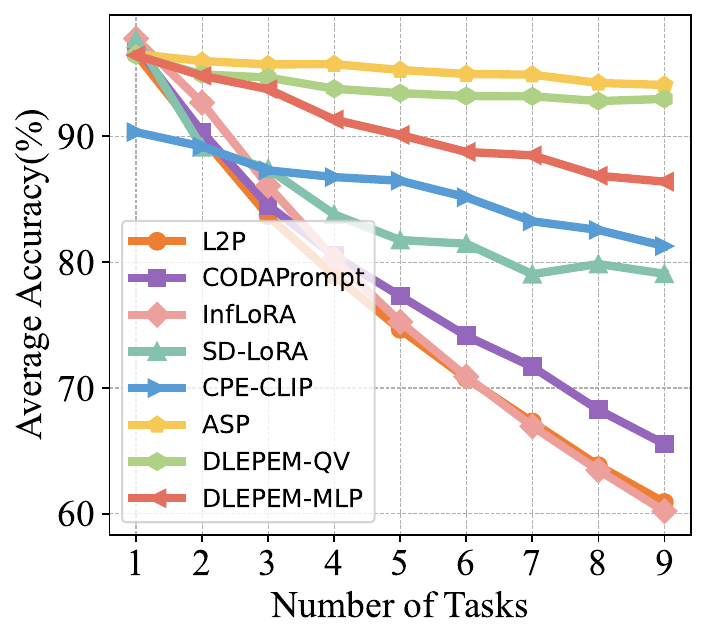}
        \end{minipage}}
    \caption{\textcolor{red}{Incremental accuracy curves on the FSCIL benchmarks. The three subplots show the CUB200, CIFAR100, and \textit{mini}ImageNet protocols, respectively, using the same ViT-B/16-IN21K backbone.}}
    \label{fig:benchmark-fscil}
\end{figure}

\textbf{Few-Shot Class-Incremental Learning:}
\textcolor{red}{We further evaluate DLEPEM in the few-shot class-incremental learning setting.} As shown in Table~\ref{tab:fscil_benchmarks}, DLEPEM consistently delivers superior accuracy and achieves leading results among the evaluated methods on multiple benchmarks. In particular, it achieves the highest last accuracy ($\mathcal{A}_L$) and average accuracy ($\bar{\mathcal{A}}$) on CUB200 and CIFAR100. For instance, on CUB200, DLEPEM-QV achieves an average accuracy of 88.77\%, outperforming the strongest baseline, ASP, by a clear margin of 5.31\%. On CIFAR100, DLEPEM-QV reaches 90.50\%, surpassing ASP by 1.96\%. While its performance on \textit{mini}ImageNet is highly competitive and on par with the strongest baselines, these substantial gains on the other datasets highlight DLEPEM's effectiveness in long-term continual learning, even under severe data sparsity. Figure~\ref{fig:benchmark-fscil} further illustrates that DLEPEM maintains superior performance throughout the incremental learning process, showing its robustness and adaptability in few-shot scenarios.

\begin{table}[H]\scriptsize
    \centering
    \caption{\textcolor{red}{Performance comparison on FSCIL benchmarks using the same ViT-B/16-IN21K backbone. Results are reported as mean \(\pm\) standard deviation over three runs; the best result in each column is highlighted in bold.}
        }
    \label{tab:fscil_benchmarks}
    \resizebox{1.0\linewidth}{!}{
        \begin{tabular}{@{}lccccccccc@{}}
            \toprule
            \multirow{2}{*}{Method} & 
            \multicolumn{3}{c}{CUB200 ($T$=11)} & 
            \multicolumn{3}{c}{CIFAR100 ($T$=9)} &
            \multicolumn{3}{c}{\textit{mini}ImageNet ($T$=9)}
            \\
            \cmidrule(lr){2-4} \cmidrule(l){5-7} \cmidrule(l){8-10}
            & $\mathcal{A}_{\text{Base}}$ & $\mathcal{A}_L$ & $\bar{\mathcal{A}}$ 
            & $\mathcal{A}_{\text{Base}}$ & $\mathcal{A}_L$ & $\bar{\mathcal{A}}$ 
            & $\mathcal{A}_{\text{Base}}$ & $\mathcal{A}_L$ & $\bar{\mathcal{A}}$\\
            \midrule
        
        L2P~\cite{wang2022learning} 
            & $91.50 \pm 0.00$ & $50.04 \pm 0.00$ & $66.70 \pm 0.00$ 
            & $93.43 \pm 0.00$ & $55.75 \pm 0.00$ & $71.81 \pm 0.00$
            & $96.53 \pm 0.00$ & $60.91 \pm 0.00$ & $76.28 \pm 0.00$
            \\
        
        CODA-Prompt~\cite{smith2023coda}
            & $91.50 \pm 0.00$ & $53.65 \pm 0.00$ & $69.30 \pm 0.00$ 
            & $94.05 \pm 0.00$ & $57.10 \pm 0.00$ & $73.11 \pm 0.00$
            & $97.15 \pm 0.00$ & $65.55 \pm 0.00$ & $78.83 \pm 0.00$
            \\
        
        InfLoRA~\cite{liang2024inflora} 
            & $92.45 \pm 0.20$ & $45.18 \pm 2.50$ & $66.27 \pm 1.43$ 
            & $\bf{94.92 \pm 0.06}$ & $57.41 \pm 0.31$ & $74.28 \pm 0.28$
            & $97.42 \pm 0.06$ & $51.52 \pm 0.10$ & $71.55 \pm 0.03$
            \\
        
        SD-LoRA~\cite{wu2025sdlora} 
            & $91.92 \pm 0.00$ & $56.28 \pm 0.00$ & $70.87 \pm 0.00$ 
            & $94.60 \pm 0.00$ & $73.51 \pm 0.00$ & $78.42 \pm 0.00$
            & $\bf{97.72 \pm 0.00}$ & $79.07 \pm 0.00$ & $84.36 \pm 0.00$
            \\
        \midrule
        CPE-CLIP~\cite{10350931}
            & $80.21 \pm 0.85$ & $63.32 \pm 0.17$ & $69.37 \pm 0.37$ 
            & $88.32 \pm 0.04$ & $79.99 \pm 0.18$ & $83.38 \pm 0.10$
            & $90.14 \pm 0.05$ & $81.55 \pm 0.04$ & $85.54 \pm 0.04$
            \\
        ASP~\cite{liu2024few}
            & $87.14 \pm 0.11$ & $82.86 \pm 0.26$ & $83.46 \pm 0.22$ 
            & $91.77 \pm 0.09$ & $86.04 \pm 0.03$ & $88.54 \pm 0.03$
            & $96.32 \pm 0.15$ & $93.72 \pm 0.25$ & $94.97 \pm 0.21$
            \\
        PriViLege~\cite{park2024pre} 
            & $82.21 \pm 0.35$ & $75.08 \pm 0.52$ & $77.50 \pm 0.33$ 
            & $90.88 \pm 0.20$ & $86.06 \pm 0.32$ & $88.08 \pm 0.20$
            & $96.68 \pm 0.06$ & $\bf{94.10 \pm 0.13}$ & $\bf{95.27 \pm 0.11}$
            \\
        \midrule
        DLEPEM-MLP 
            & $92.51 \pm 0.00$ & $85.37 \pm 0.00$ & $88.53 \pm 0.00$ 
            & $94.28 \pm 0.00$ & $84.67 \pm 0.00$ & $89.00 \pm 0.00$
            & $96.37 \pm 0.00$ & $89.96 \pm 0.00$ & $93.11 \pm 0.00$
            \\
        DLEPEM-QV 
            & $\bf{92.68 \pm 0.00}$ & $\bf{86.22 \pm 0.00}$ & $\bf{88.77 \pm 0.00}$ 
            & $94.00 \pm 0.00$ & $\bf{87.29 \pm 0.00}$ & $\bf{90.50 \pm 0.00}$
            & $96.77 \pm 0.00$ & $93.62 \pm 0.00$ & $94.80 \pm 0.00$
            \\
            \bottomrule
        \end{tabular}
        }
\end{table}

\begin{table}[H]\scriptsize
    \centering
    \caption{\textcolor{red}{Ablation study on CIL and FSCIL tasks. The first two benchmarks follow CIL protocols, while the last two follow FSCIL protocols. Each metric reports MLP-Expert / QV-Expert performance.}}
    \label{tab: ablation_studies_components_cil}
    \resizebox{1.0\linewidth}{!}{
        \begin{tabular}{@{}l|cccc|cccc@{}}
            \toprule
            \multirow{2}{*}{Ablated Components} 
            & \multicolumn{2}{c}{CIFAR100 ($T$=10)} 
            & \multicolumn{2}{c|}{ImageNet-R ($T$=5)} 
            & \multicolumn{2}{c}{CIFAR100 ($T$=9)} 
            & \multicolumn{2}{c}{\textit{mini}ImageNet ($T$=9)}\\
            \cmidrule(lr){2-3} \cmidrule(lr){4-5} \cmidrule(lr){6-7}\cmidrule(lr){8-9}
            & $\mathcal{A}_L$ & $\bar{\mathcal{A}}$
            & $\mathcal{A}_L$ & $\bar{\mathcal{A}}$
            & $\mathcal{A}_L$ & $\bar{\mathcal{A}}$
            & $\mathcal{A}_L$ & $\bar{\mathcal{A}}$\\
            \midrule
            \textcolor{red}{w/o Dynamic LoRA-Expert}
            &83.13\ /\ 86.72 &88.85\ /\ 90.82 
            &61.17\ /\ 72.37 &74.31\ /\ 80.14
            &73.03\ /\ 79.73 &80.39\ /\ 86.59  
            &84.26\ /\ 91.11&89.90\ /\ 93.31
            \\
            w/o Prototype-Ensemble 
            &85.79\ /\ 87.59 &91.21\ /\ 92.29 
            &69.53\ /\ 73.13 &77.75\ /\ 80.42
            &81.13\ /\ 84.09 & 86.97\ /\ 88.67 
            &85.54\ /\ 92.25 &90.03\ /\ 94.06
            \\
            \midrule
            DLEPEM-MLP\ /\ QV 
            &\bf{85.99\ /\ 88.84} &\bf{92.40\ /\ 93.39} 
            &\bf{76.25\ /\ 78.77} &\bf{82.37\ /\ 83.43} 
            &\bf84.67\ /\ 87.29 &\bf89.00\ /\ 90.5 
            &\bf89.96\ /\ 93.62&\bf93.11\ /\ 94.80\\
            \bottomrule
        \end{tabular} 
        }
\end{table}

\subsection{Ablation Study}
\textbf{Different Components:} 
\textcolor{red}{We conduct ablation studies to assess the contribution of each component in DLEPEM (Table~\ref{tab: ablation_studies_components_cil}). The variant \textbf{w/o Dynamic LoRA-Expert} uses a single LoRA-Expert for all tasks, resulting in a significant performance drop under large domain shifts and indicating severe forgetting and task interference. In contrast, assigning a dedicated LoRA-Expert to each task better preserves the stability-plasticity trade-off across incremental steps. This result underscores the value of task-specific experts in PEFT-based continual learning. Removing the prototype-ensemble router (\textbf{w/o Prototype-Ensemble}) and using frozen class prototypes as keys also degrades performance, confirming the router's critical role in effective module-sample matching.}
\textcolor{red}{The advantage of dynamic expert allocation is most visible on ImageNet-R. Replacing the single shared expert with task-specific dynamic experts increases DLEPEM-MLP from 61.17/74.31 to 76.25/82.37 in $\mathcal{A}_L/\bar{\mathcal{A}}$, and increases DLEPEM-QV from 72.37/80.14 to 78.77/83.43. These gains support our claim that isolating LoRA parameters by task reduces cross-task interference, especially under domain shift.}

\textbf{Different Routers:} \textcolor{red}{In addition to the prototype-ensemble strategy, we compare three module-sample matching mechanisms: frozen-PTM prototypes $\mathbf{P}^F$, K-nearest-neighbor (KNN) matching, and router-domain prototypes from $\mathbf{E}^{router}$. For KNN, features are extracted using the frozen PTM, and performance is evaluated across $k$ values ($k=\{1,3,5,7,9\}$), with $k=3$ yielding the best results. Both $\mathbf{P}^F$ and $\mathbf{E}^{router}$ provide single-source keys within the prototype-ensemble framework.} 

\begin{figure}[t]
    \begin{center}
        {\includegraphics[width=0.7\linewidth]{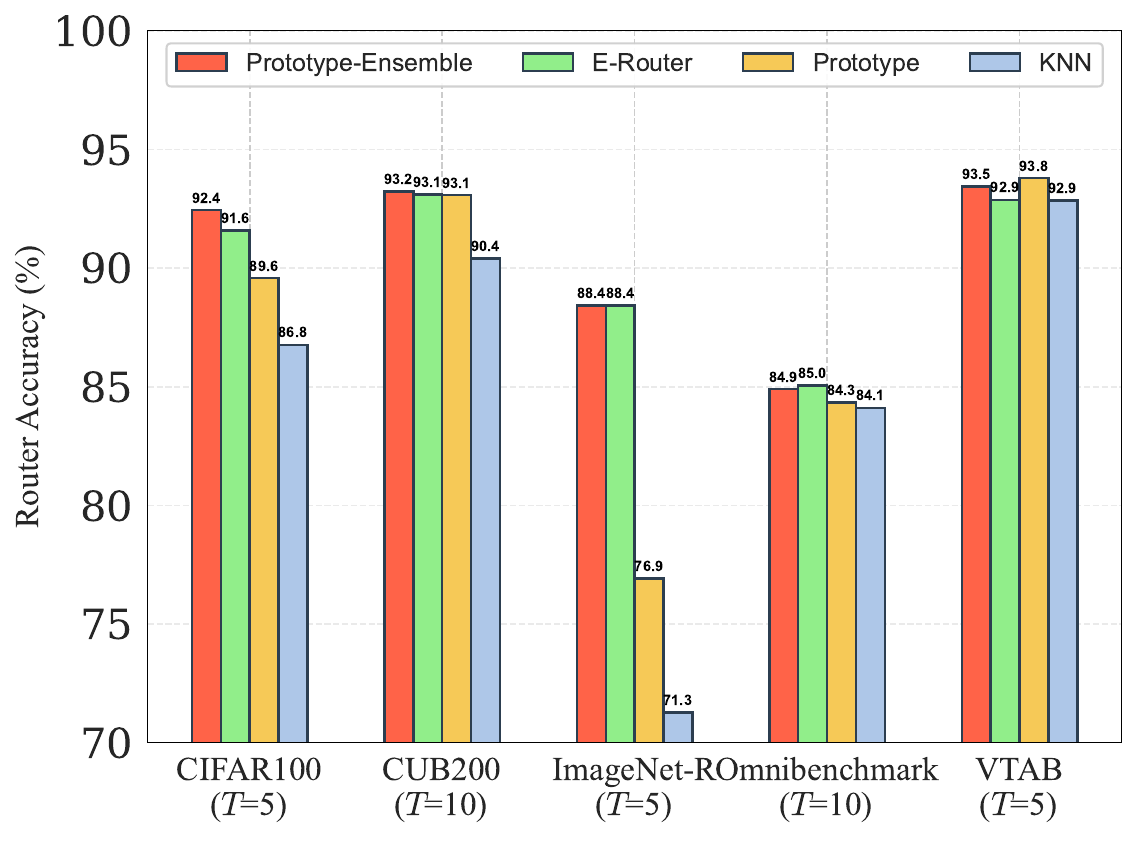}}
    \end{center}
    \caption{\textcolor{red}{Comparison of routing strategies for module-identity inference. The prototype-ensemble strategy is compared with frozen-PTM prototypes, KNN matching, and router-domain prototypes under the same evaluation protocols.}}
    \label{fig:router_compare}
\end{figure}

\begin{table}[H]\scriptsize
    \centering
    \caption{\textcolor{red}{Diagnostic accuracy of the classifier, router, and oracle expert selection on five CIL benchmarks. Each entry is reported as DLEPEM-MLP / DLEPEM-QV; the oracle expert setting uses the ground-truth task identity only for diagnostic analysis.}}
    \label{tab:cnn-router-lora-expert}
    \resizebox{\linewidth}{!}{
        \begin{tabular}{lccccc}
            \toprule
            Metric & CIFAR100 ($T=10$) & CUB200 ($T=10$) & ImageNet-R ($T=5$) & OmniBenchmark ($T=10$) & VTAB ($T=5$) \\
            \midrule
            \textcolor{red}{CNN}
            & \textcolor{red}{$92.40\ /\ 93.39$}
            & \textcolor{red}{$92.31\ /\ 92.09$}
            & \textcolor{red}{$82.37\ /\ 83.43$}
            & \textcolor{red}{$81.60\ /\ 82.16$}
            & \textcolor{red}{$91.84\ /\ 91.11$} \\
            \textcolor{red}{Router}
            & \textcolor{red}{$91.88\ /\ 93.68$}
            & \textcolor{red}{$93.24\ /\ 93.31$}
            & \textcolor{red}{$88.43\ /\ 88.87$}
            & \textcolor{red}{$84.90\ /\ 85.55$}
            & \textcolor{red}{$93.45\ /\ 93.56$} \\
            \textcolor{red}{LoRA-Expert (Oracle)}
            & \textcolor{red}{$98.04\ /\ 97.81$}
            & \textcolor{red}{$97.15\ /\ 96.31$}
            & \textcolor{red}{$89.17\ /\ 88.81$}
            & \textcolor{red}{$92.88\ /\ 91.48$}
            & \textcolor{red}{$97.66\ /\ 95.88$} \\
            \bottomrule
        \end{tabular}
    }
\end{table}

\textcolor{red}{In Table~\ref{tab:cnn-router-lora-expert}, each entry is reported as DLEPEM-MLP / DLEPEM-QV. CNN reports the final average classification accuracy of each variant. Router reports the expert-selection accuracy of the learned routing module. LoRA-Expert (Oracle) reports classification accuracy when the ground-truth task/expert identity is used at test time to select the correct LoRA-Expert before normal within-expert classification.}

As shown in Figure~\ref{fig:router_compare}, the prototype ensemble consistently outperforms the baseline KNN approach. Two key insights emerge:

\textbf{1. Domain-specific adaptation:} Under significant domain shift (e.g., ImageNet-R $\left(T=5\right)$), the ensemble achieves larger gains, primarily due to $\mathbf{E}^{router}$'s ability to capture domain-specific characteristics.

\textbf{2. Robust matching via integration:} Combining generalized and domain-aware prototypes enables more reliable module-sample matching. Even when $\mathbf{E}^{router}$ underperforms $\mathbf{P}^F$, their ensemble compensates through mutual alignment, enhancing overall stability and accuracy.
\textcolor{red}{This conclusion is also consistent with the component ablation in Table~\ref{tab: ablation_studies_components_cil}: on ImageNet-R, replacing prototype-ensemble retrieval with frozen-prototype keys reduces $\mathcal{A}_L/\bar{\mathcal{A}}$ from 76.25/82.37 to 69.53/77.75 for DLEPEM-MLP and from 78.77/83.43 to 73.13/80.42 for DLEPEM-QV. The result indicates that combining PTM-general and router-domain prototypes provides more reliable MII than relying on a single feature source.}
\textcolor{red}{Figure~\ref{fig:router_compare} also contains the two single-source prototype variants requested by the reviewer: the frozen-PTM prototype \(\mathbf{P}^{F}\) and the router-domain prototype \(\mathbf{E}^{router}\). Across the evaluated settings, both single-source variants, including the router-domain prototype variant, obtain lower accuracy than the prototype-ensemble strategy, which supports the complementarity of the two prototype sources.}
\textcolor{red}{This router comparison also serves as a post-hoc compatibility check for the append-only dictionary. During evaluation, current queries are matched against stored keys accumulated from previous stages. If historical router-domain keys were incompatible with the current router state, prototype-ensemble retrieval would not consistently outperform PTM-only or KNN-style matching. The observed gains therefore suggest that the frozen-PTM component and stability-regularized router updates help maintain usable key-query compatibility across incremental stages.}

\begin{figure}[t]
    \centering
    \subfloat[ImageNet-R ($T=10$).\label{fig:ablation_study_ptm_inr}]{%
        \begin{minipage}{0.49\linewidth}
        \centering
        \includegraphics[width=\linewidth]{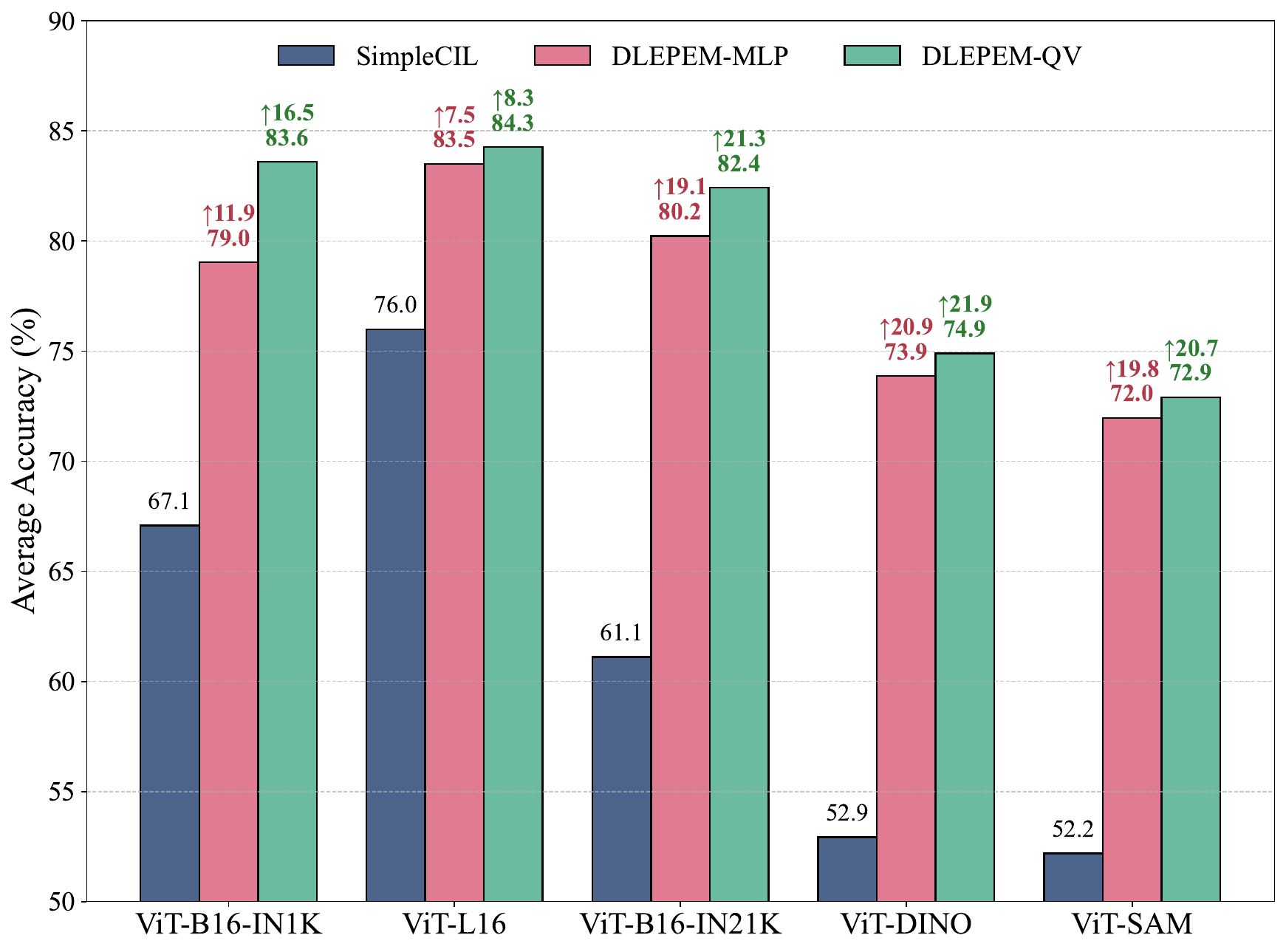}
        \end{minipage}}
    \subfloat[CIFAR100 ($T$=10).\label{fig:ablation_study_ptm_cifar}]{%
        \begin{minipage}{0.49\linewidth}
        \centering
        \includegraphics[width=\linewidth]{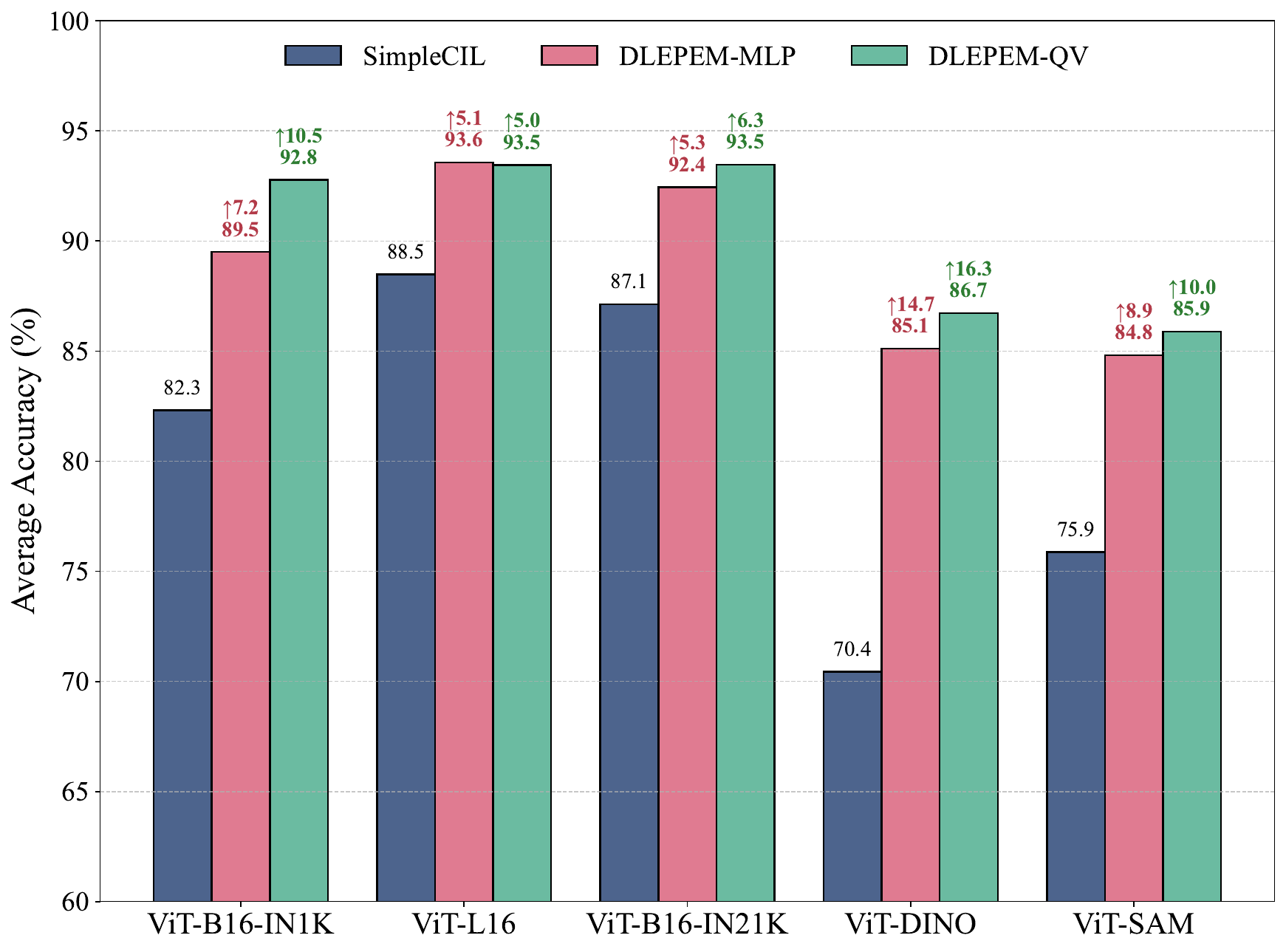}
        \end{minipage}}
    \caption{\textcolor{red}{Effect of different pre-trained backbones on CIL performance. The comparison evaluates DLEPEM with several ViT-based pre-training sources and reports the corresponding performance on ImageNet-R and CIFAR100.}}
    \label{fig:ablation-study-settings}
\end{figure}

\textcolor{red}{\textbf{Different Pre-trained Models:}} 
Beyond validating DLEPEM on ViT-B/16-IN21K, we further assess its generalization across diverse Transformer-based architectures, including ViT-B/16-IN1K \& ViT-L/16~\cite{dosovitskiy2020image}, ViT-B/16-DINO~\cite{caron2021emerging}, and ViT-B/16-SAM~\cite{chen2021vision}. As a baseline, SimpleCIL~\cite{zhou2023revisitingclassincrementallearningpretrained} fine-tunes only the classifier to reflect the inherent capability of each backbone in incremental settings.
\textcolor{red}{We evaluate DLEPEM on the CIFAR100 and ImageNet-R benchmarks, with results shown in Figure~\ref{fig:ablation-study-settings}(a) and (b). Two key observations emerge:}

1. DLEPEM yields greater improvements on larger models (e.g., ViT-L/16), highlighting its scalability with model capacity.

2. Among similarly sized architectures (e.g., ViT-B/16 variants), DLEPEM consistently outperforms the SimpleCIL baseline, demonstrating robustness to architectural and pretraining differences.

These results show DLEPEM's versatility and effectiveness across a wide range of Transformer-based backbones.




\subsection{Analysis of Trainable Parameters and Accuracy} 
\begin{figure}[t]
    \centering
    \subfloat[CIFAR100 ($T$=10)\label{fig:parameters_acc_compare_cifar}]{%
        \begin{minipage}{0.49\linewidth}
        \centering
        \includegraphics[width=\linewidth]{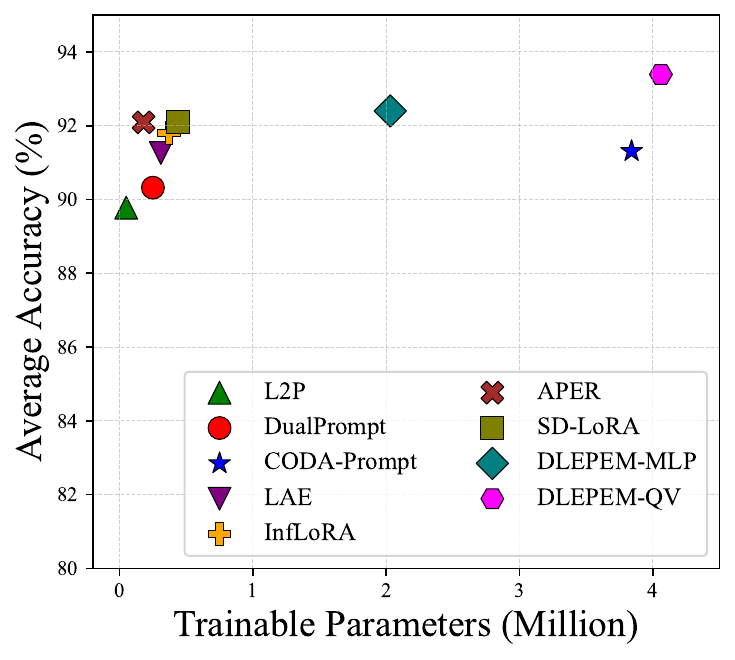}
        \end{minipage}}
    \subfloat[VTAB ($T$=5)\label{fig:parameters_acc_compare_vtab}]{%
        \begin{minipage}{0.49\linewidth}
        \centering
        \includegraphics[width=\linewidth]{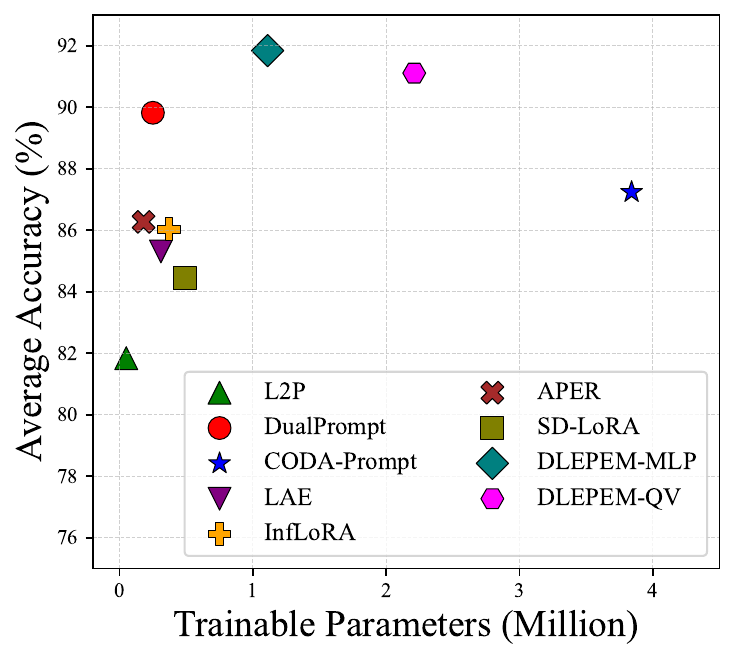}
        \end{minipage}}
    \caption{\textcolor{red}{Parameter--accuracy comparison on CIL benchmarks. The plots compare trainable parameter counts and final accuracy on CIFAR100 and VTAB to illustrate the parameter-performance trade-off of DLEPEM.}}
    \label{fig:parameters_acc_compare}
\end{figure}

%
\textcolor{red}{To further evaluate the parameter-performance trade-off, we compare the number of trainable parameters with accuracy across different methods in Figure~\ref{fig:parameters_acc_compare}. As illustrated in these plots, DLEPEM achieves a favorable balance between model size and performance, demonstrating that it uses additional parameters efficiently to improve accuracy.}

\textcolor{red}{We further quantify the scalability of the one-expert-per-task design. DLEPEM is rehearsal-free in the sense that it stores no raw samples, old feature batches, or exemplar buffers; nevertheless, it retains a frozen LoRA-Expert bank, the current router, and an append-only prototype dictionary. For a ViT with hidden dimension $d$, LoRA rank $r$, $L$ Transformer blocks, and $b$ bytes per floating-point value, the storage of one MLP-Expert, one QV-Expert, and the prototype dictionary after task $t$ can be written as}
\begin{equation}
\textcolor{red}{
\begin{aligned}
N_{\mathrm{MLP}} &= L(2dr),\\
N_{\mathrm{QV}} &= 2L(2dr),\\
M_{\mathrm{proto}}(t) &= 2d|\mathcal{Y}_{1:t}|b .
\end{aligned}}
\label{eq:storage_scalability}
\end{equation}
\textcolor{red}{For the LoRA-Expert bank, under our default ViT-B/16-IN21K setting, $d=768$, $L=12$, $r=10$, and $b=4$ for FP32. Thus, each MLP-Expert adds $0.18$M parameters ($\approx 0.70$ MiB), and each QV-Expert adds $0.37$M parameters ($\approx 1.41$ MiB). In a 10-task setting, the frozen expert bank stores $1.84$M LoRA parameters for DLEPEM-MLP and $3.69$M for DLEPEM-QV, corresponding to approximately $2.1\%$ and $4.3\%$ of an 86M-parameter ViT-B backbone, respectively. For longer task streams, the growth is linear: in a 50-task sequence, the expert bank contains $9.22$M parameters for DLEPEM-MLP and $18.43$M for DLEPEM-QV; in a 100-task sequence, it contains $18.43$M and $36.86$M parameters, respectively. The router contributes one additional same-size LoRA module and is constant with respect to the number of tasks. Therefore, the LoRA-Expert bank remains lightweight under the evaluated protocols, but it is the main component that scales with the number of tasks.}

\textcolor{red}{For the prototype dictionary, each ensemble prototype contains two 768-dimensional vectors: one general prototype from the frozen PTM feature space and one domain-specific prototype from the router feature space. Its memory therefore scales with the number of seen classes rather than directly with the number of tasks. In FP32, the dictionary requires about $0.59$ MiB for 100 classes, $1.17$ MiB for 200 classes, and $1.76$ MiB for 300 classes. Compared with both the LoRA-Expert bank and the ViT-B backbone, this storage is small in our evaluated benchmarks, but it still grows as $O(|\mathcal{Y}_{1:T}|)$ with the number of seen classes.}

\subsection{Training and Inference Time Comparison}
\textcolor{red}{At inference time, task-specific LoRA-Expert selection is based on nearest-neighbor search using cosine similarity, formulated as \( \hat{i} = \arg\max_{i \in \mathbf{Y}_t} \mathrm{cos}(\mathbf{q}(\mathbf{x}), \mathbf{K}_i) \). Table~\ref{tab:time_comparison} reports training time and inference latency. The reported inference latency is measured end-to-end for one image and includes query construction with the frozen PTM and router, nearest-neighbor expert selection in the prototype dictionary, the selected LoRA-Expert forward pass, and final prototype-weight classification. DLEPEM-MLP requires $69.46$ s/epoch on CIFAR100 and $33.30$ s/epoch on ImageNet-R, which is competitive with representative PEFT-based baselines. Its inference latency is $8.39$ ms/image on CIFAR100 and $8.47$ ms/image on ImageNet-R. This latency is higher than that of SD-LoRA and InfLoRA, so the overhead should not be described as negligible. A more accurate interpretation is that DLEPEM introduces a non-negligible but still practical inference overhead in exchange for stronger expert retrieval and final accuracy.}

\textcolor{red}{From a deployment-memory perspective, the storage analysis above shows that DLEPEM remains lightweight relative to the ViT-B/16-IN21K backbone under the evaluated protocols. For the LoRA-Expert bank, each MLP-Expert adds about $0.70$ MiB and each QV-Expert adds about $1.41$ MiB in FP32; in a 10-task setting, the retained expert bank accounts for about $2.1\%$ and $4.3\%$ of the 86M-parameter backbone for DLEPEM-MLP and DLEPEM-QV, respectively. For the prototype dictionary, the storage is smaller and requires at most about $1.76$ MiB for 300 seen classes in our evaluated CIL benchmarks. Therefore, DLEPEM is suitable for GPU-based real-time or near-real-time recognition scenarios, while stricter edge deployment or much longer task streams may require additional expert compression or pruning.}

\begin{table}[H]
    \centering
    \caption{\textcolor{red}{Training and inference time comparison. Training time is the average time per epoch for each incremental task, and inference latency is measured in milliseconds per image. All methods use the same ViT-B/16-IN21K backbone for a fair comparison.}}
    \label{tab:time_comparison}
    \resizebox{1.0\linewidth}{!}{
        \begin{tabular}{@{}lcccc@{}}
            \toprule
            \multirow{2}{*}{Method} & 
            \multicolumn{2}{c}{CIFAR100 ($T$=10)} & 
            \multicolumn{2}{c}{ImageNet-R ($T$=5)} \\
            \cmidrule(lr){2-3} \cmidrule(l){4-5} 
            & Training Time(s) & Inference Time(ms) & Training Time(s) & Inference Time(ms)\\
            \midrule
            L2P~\cite{wang2022learning}          &102.12&3.77  &50.12 &3.84 \\
            DualPrompt~\cite{wang2022dualprompt}    &93.21 &3.44  &45.16 &3.58 \\
            CODA-Prompt~\cite{smith2023coda}    &99.42 &2.99  &47.53 &3.08 \\
            LAE~\cite{gao2023unified}           &46.87 &3.35  &24.26 &3.52 \\
            InfLoRA~\cite{liang2024inflora}        &72.36 &2.03  &34.96 &2.24 \\
            APER~\cite{zhou2023revisitingclassincrementallearningpretrained}  &16.04 &3.63  &8.92 &3.71 \\
            SD-LoRA~\cite{wu2025sdlora}  &79.08&1.93 &32.85&2.04 \\
            \midrule
            DLEPEM-MLP \ /\ \color{red}{QV}    &69.46 \ /\  \color{red}{71.37} &8.39 \ /\ \color{red}{9.26}   &33.3 \ /\ \color{red}{34.15} &8.47 \ /\ \color{red}{9.08} \\
            \bottomrule
        \end{tabular}}
        
\end{table}

\subsection{Parameter Sensitivity Analysis}

We investigate the sensitivity of three key hyperparameters in DLEPEM: (1) the LoRA rank $r$, (2) the insertion positions of LoRA-Experts in the ViT backbone, and (3) the distillation coefficient $\alpha$ for plasticity regularization.

\begin{figure}[t!]
    \centering
    \subfloat[\textcolor{red}{Rank $r$ and LoRA insertion position.}\label{fig:hyperparameters}]{%
        \begin{minipage}{0.47\linewidth}
        \centering
        \includegraphics[width=\linewidth]{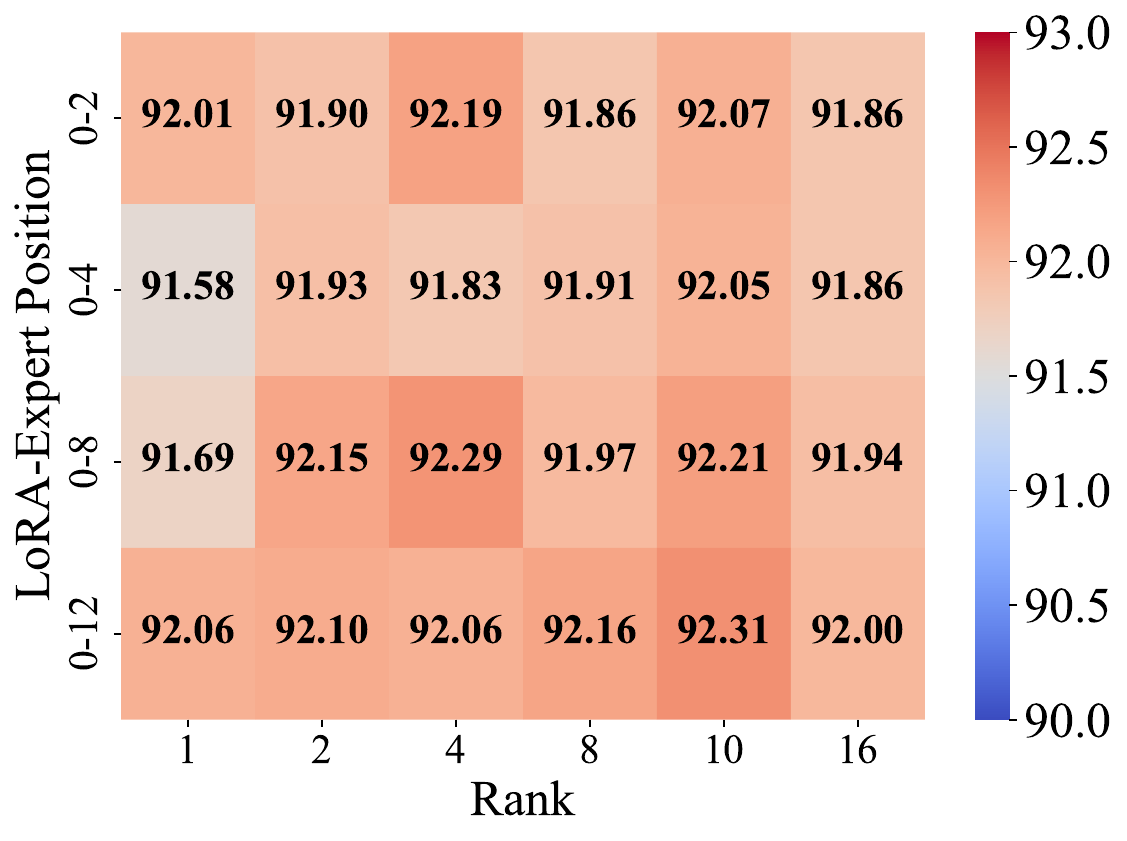}
        \end{minipage}}
    \subfloat[\textcolor{red}{Distillation coefficient $\alpha$.}\label{fig:parameter_analysis_alpha.png}]{%
        \begin{minipage}{0.47\linewidth}
        \centering
        \includegraphics[width=\linewidth]{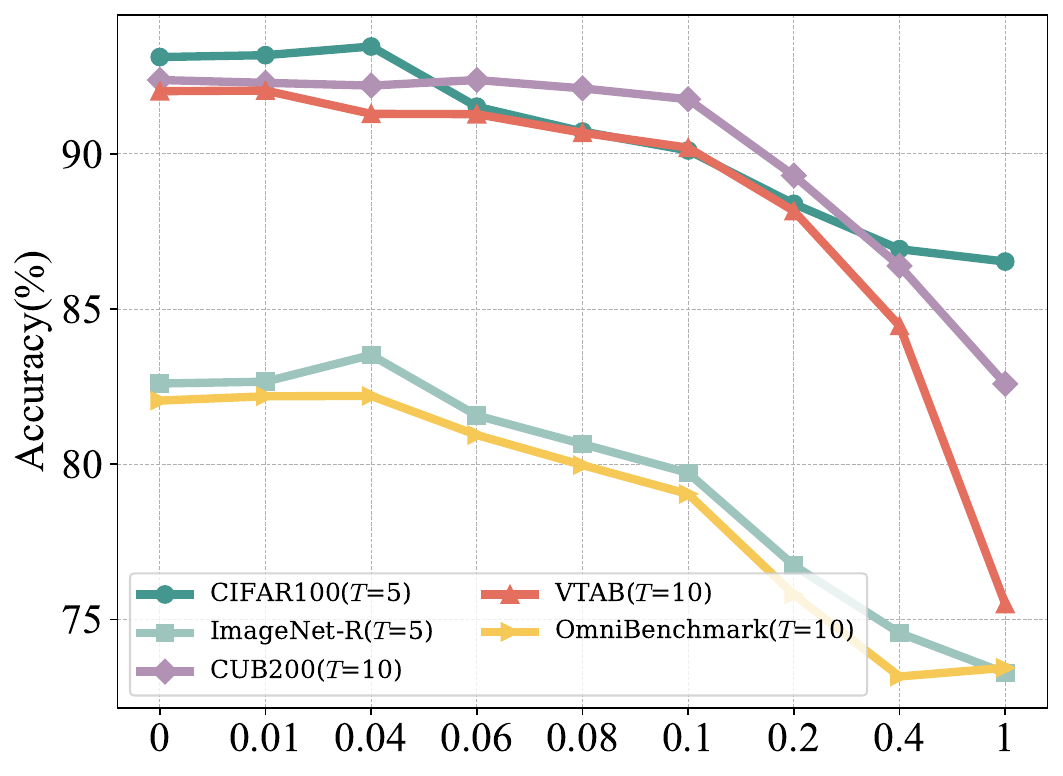}
        \end{minipage}}
    \caption{\textcolor{red}{Hyperparameter sensitivity of DLEPEM. Subplot (a) analyzes the effect of LoRA rank and insertion position, while subplot (b) shows the effect of the distillation coefficient \(\alpha\) across datasets.}}
    \label{fig:parameters_compare}
\end{figure}

To evaluate the effect of $r$ and insertion positions, we conduct experiments on CUB200 ($T=10$), varying
$r \in \{1, 2, 4, 8, 10, 16\}$ and testing insertion ranges \{0-2, 0-4, 0-8, 0-12\}, where ``0-2'' denotes insertion into the first two Transformer layers. As shown in Figure~\ref{fig:parameters_compare}(a), DLEPEM maintains stable performance across various settings, demonstrating robustness to hyperparameter variations. Following prior work~\cite{liang2024inflora,wu2025sdlora}, we adopt $r=10$ and insert LoRA-Experts into all Transformer blocks. Similar trends are observed on other datasets.

We also analyze the impact of the distillation coefficient $\alpha$ across datasets (Figure~\ref{fig:parameters_compare}(b)). \textcolor{red}{The two endpoints correspond to removing one distillation term: \(\alpha=0\) removes plasticity feature distillation and keeps only \(\mathcal{L}_{SFD}\), while \(\alpha=1\) removes stability feature distillation and keeps only \(\mathcal{L}_{PFD}\).} The model performs consistently well for $\alpha \in (0,0.1]$, with $\alpha=0.04$ as the default.

\subsection{Performance of LoRA-based methods across sequential tasks}
\label{sec:old-new-analysis}
\textcolor{red}{Kim and Han~\cite{kim2023stabilityplasticity} show that final or average accuracy can obscure whether a CIL method is genuinely plastic or mainly stable, and they propose feature-representation diagnostics such as classifier retraining with frozen feature extractors and representation-similarity analysis. We do not reproduce that full representation-level protocol here. Instead, using the available sequential-task results, we provide a behavior-level decomposition: new-task performance is used as an empirical proxy for plasticity, and old-task performance after subsequent learning is used as an empirical proxy for stability.}

\begin{figure}[t]
    \centering
    \subfloat[New task performance.\label{fig:lora_new_acc}]{%
        \begin{minipage}{0.49\linewidth}
        \centering
        \includegraphics[width=0.98\linewidth]{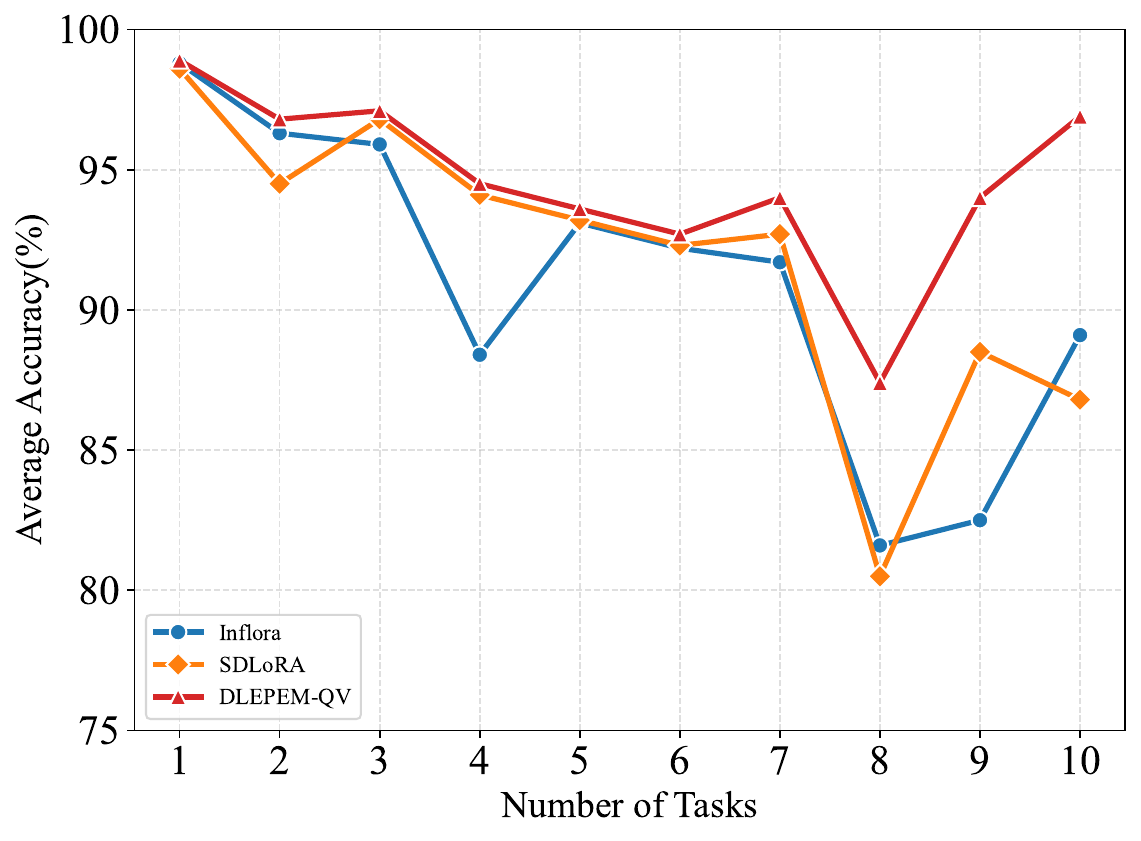}
        \end{minipage}}
    \subfloat[Old task performance.\label{fig:lora_old_acc}]{%
        \begin{minipage}{0.49\linewidth}
        \centering
        \includegraphics[width=0.98\linewidth]{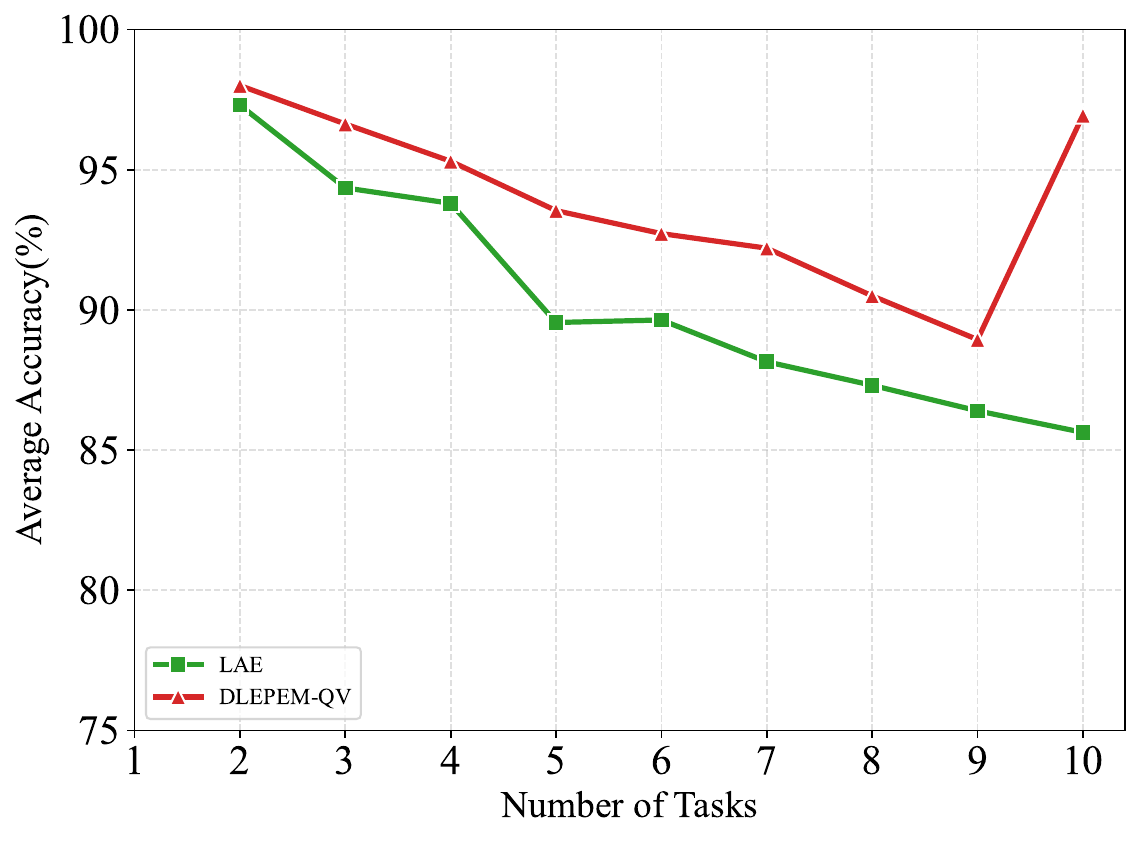}
        \end{minipage}}
    \caption{\textcolor{red}{New-task and old-task performance of LoRA-based CIL methods. The left subplot reports accuracy on newly introduced tasks as a proxy for plasticity, and the right subplot reports retained accuracy on previous tasks as a proxy for stability.}}
    \label{fig:performance_new_old_acc}
\end{figure}

\textcolor{red}{To further analyze the performance characteristics of LoRA-based methods, we evaluate LAE~\cite{gao2023unified}, InfLoRA~\cite{liang2024inflora}, and SD-LoRA~\cite{wu2025sdlora} in terms of model plasticity and stability on CIFAR100 ($T=10$). As shown in Figure~\ref{fig:performance_new_old_acc}(a), both InfLoRA and SD-LoRA show lower performance on new tasks because their gradient-direction constraints are designed to preserve old knowledge. While these constraints effectively mitigate catastrophic forgetting, they can compromise the model's plasticity when learning new tasks. In contrast, DLEPEM learns task-specific LoRA modules without such restrictions, enabling stronger adaptation to new tasks while maintaining competitive performance.}
\textcolor{red}{This comparison further distinguishes DLEPEM from constraint-based LoRA continual learning: InfLoRA and SD-LoRA protect old knowledge by restricting update directions or magnitudes, whereas DLEPEM preserves old knowledge structurally by freezing previous experts while keeping the current expert fully trainable within its low-rank subspace. Thus, DLEPEM shifts the stability-plasticity trade-off from a single shared update space to two separated factors: parameter-stationary old experts for stability and a newly trainable expert for plasticity.}

LAE employs a different strategy by continuously integrating new parameters into the existing model through weighted blending. However, as demonstrated in Figure~\ref{fig:performance_new_old_acc}(b), this approach leads to progressive destabilization of previously learned knowledge. DLEPEM consistently outperforms LAE on old tasks, demonstrating that our dynamic expert selection mechanism effectively preserves old knowledge without the instability issues associated with parameter blending.

\section{Conclusion}
We propose DLEPEM, a novel framework for CIL that combines Dynamic LoRA-Experts for task-specific adaptation and Prototype-Ensemble Matching for improved module selection.
\textcolor{red}{This dual design supports a more favorable empirical stability-plasticity trade-off under the evaluated CIL protocols by combining parameter-stationary old experts for stability with a trainable expert for each new task to preserve plasticity.}
Extensive evaluations on six benchmarks demonstrate that DLEPEM achieves strong and competitive performance under the evaluated protocols. \textcolor{red}{Although each LoRA-Expert is lightweight, the stored expert bank grows linearly with the number of tasks, and the prototype dictionary grows with the number of seen classes. Future work will explore expert compression, rank sharing, expert pruning, expert merging/distillation, different prototype fusion weights, random and expert-ID routing diagnostics, and extensions to multi-modal learning.}

\begin{table}[H]
\caption{\textcolor{red}{Notation used in DLEPEM.}}
\label{tab:notation}
\centering
\begingroup
\color{black}
\begin{tabular}{p{0.22\linewidth}p{0.68\linewidth}}
\hline
Symbol & Definition \\
\hline
\(\mathcal{D}_t\), \(\mathcal{Y}_t\) & Training set and class set of task \(t\). \\
\(\mathbf{Y}_t\) & The set of all classes observed up to task \(t\), i.e., \(\mathbf{Y}_t=\mathcal{Y}_1\cup\cdots\cup\mathcal{Y}_t\). \\
\(\mathbf{E}_t\) & Task-\(t\) LoRA-Expert module, i.e., a group of low-rank LoRA adapters inserted into selected Transformer layers rather than a complete independent ViT. Historical experts are frozen after training. \\
\(\mathbf{U}_t\), \(\mathbf{V}_t\), \(r\) & Low-rank matrices and rank inside each adapter. For row-vector input \(\mathbf{e}\in\mathbf{R}^{1\times d_{in}}\), \(\mathbf{U}_t\in\mathbf{R}^{d_{in}\times r}\), \(\mathbf{V}_t\in\mathbf{R}^{r\times d_{out}}\), with \(r=10\) by default. \\
\(\mathbf{V}_{attn}\) & Attention value matrix in Eq.~(\ref{eq:attn}), distinct from the LoRA matrix \(\mathbf{V}_t\). \\
\(\mathbf{E}^{router}_t\) & Router after learning task \(t\), used to extract router-domain features. \\
\(\phi(\mathbf{x})\), \(\phi(\mathbf{x};\mathbf{E})\) & Frozen-PTM feature and the feature obtained with module \(\mathbf{E}\), respectively. \\
\(\mathbf{P}^{L}_{i,t}\) & LoRA-Expert prototype of class \(i\) at task \(t\). \\
\(\mathbf{P}^{F}_{i}\), \(\mathbf{P}^{R}_{i}\) & Frozen-PTM prototype vector and router-domain prototype vector of class \(i\). \\
\(\mathbf{P}_i\) & Ensemble prototype vector of class \(i\), defined as \(\mathbf{P}_i=\mathrm{concat}(\mathbf{P}^{F}_{i},\mathbf{P}^{R}_{i})=[\mathbf{P}^{F}_{i};\mathbf{P}^{R}_{i}]\). \\
\([\cdot;\cdot]\) & Vector concatenation operator; it does not denote matrix summation. \\
\(\mathbf{K}_i\), \(\nu_i\) & Dictionary key of class \(i\) and the associated expert index. In our implementation, \(\mathbf{K}_i=\mathbf{P}_i\). \\
\(\mathbf{q}(\mathbf{x})\) & Ensemble query of test sample \(\mathbf{x}\). \\
\(\sigma(\cdot)\), \(\tau\) & Softmax shorthand and temperature used in the distillation losses, where \(\sigma(\mathbf{z})=\mathrm{softmax}(\mathbf{z})\) and \(\sigma(\mathbf{z}/\tau)=\mathrm{softmax}(\mathbf{z}/\tau)\). \\
\hline
\end{tabular}
\endgroup
\end{table}


\vspace{6pt}

\authorcontributions{Conceptualization, Hongwei Zhao; methodology, Hongwei Zhao; software, Hongwei Zhao; validation, Hongwei Zhao, Rui Liu and Yansong Liu; formal analysis, Hongwei Zhao; investigation, Hongwei Zhao; resources, Rui Liu; data curation, Hongwei Zhao; writing--original draft preparation, Hongwei Zhao; writing--review and editing, Hongwei Zhao, Rui Liu and Yansong Liu; visualization, Hongwei Zhao; supervision, Yansong Liu; project administration, Rui Liu; funding acquisition, Rui Liu. All authors have read and agreed to the published version of the manuscript.}

\institutionalreview{Not applicable.}

\informedconsent{Not applicable.}

\dataavailability{The datasets used in this study are publicly available benchmark datasets. The code is available at \url{https://github.com/hongwei-zhao/Applied_Sciences-DLEPEM-main}.}


\conflictsofinterest{The authors declare no conflicts of interest.}

\abbreviations{Abbreviations}{
The following abbreviations are used in this manuscript:
\\

\noindent
\begin{tabular}{@{}ll}
CIL & Class-Incremental Learning\\
FSCIL & Few-Shot Class-Incremental Learning\\
PEFT & Parameter-Efficient Fine-Tuning\\
PTM & Pre-Trained Model\\
LoRA & Low-Rank Adaptation\\
MoE & Mixture-of-Experts\\
ViT & Vision Transformer\\
MHA & Multi-Head Self-Attention\\
FFN & Feed-Forward Network\\
WTP & Within-Task Prediction\\
MII & Module Identity Inference
\end{tabular}
}

\begin{adjustwidth}{-\extralength}{0cm}
\reftitle{References}

\PublishersNote{}
\end{adjustwidth}


\begin{thebibliography}{999}

\bibitem[McCloskey and Cohen(1989)]{mccloskey1989catastrophic}
McCloskey, M.; Cohen, N.J.
\newblock Catastrophic Interference in Connectionist Networks: The Sequential Learning Problem. In {\em Psychology of Learning and Motivation}; Academic Press,  1989; Vol.~24, pp. 109--165.
\newblock {\url{https://doi.org/10.1016/S0079-7421(08)60536-8}}.

\bibitem[French(1999)]{french1999catastrophic}
French, R.M.
\newblock Catastrophic forgetting in connectionist networks.
\newblock {\em Trends in Cognitive Sciences} {\bf 1999}, {\em 3},~128--135.
\newblock {\url{https://doi.org/10.1016/S1364-6613(99)01294-2}}.

\bibitem[Grossberg(2012)]{grossberg2012studies}
Grossberg, S.T.
\newblock {\em Studies of mind and brain: Neural principles of learning, perception, development, cognition, and motor control}; Vol.~70, Springer Science \& Business Media,  2012.
\newblock {\url{https://doi.org/10.1007/978-94-009-7758-7}}.

\bibitem[Liang and Li(2024)]{liang2024inflora}
Liang, Y.S.; Li, W.J.
\newblock InfLoRA: Interference-Free Low-Rank Adaptation for Continual Learning.
\newblock In Proceedings of the 2024 IEEE/CVF Conference on Computer Vision and Pattern Recognition (CVPR),  2024, pp. 23638--23647.
\newblock {\url{https://doi.org/10.1109/CVPR52733.2024.02231}}.

\bibitem[Han et~al.(2021)Han, Zhang, Ding, Gu, Liu, Huo, Qiu, Yao, Zhang, Zhang, et~al.]{han2021pre}
Han, X.; Zhang, Z.; Ding, N.; Gu, Y.; Liu, X.; Huo, Y.; Qiu, J.; Yao, Y.; Zhang, A.; Zhang, L.;  et~al.
\newblock Pre-trained models: Past, present and future.
\newblock {\em AI Open} {\bf 2021}, {\em 2},~225--250.
\newblock {\url{https://doi.org/10.1016/j.aiopen.2021.08.002}}.

\bibitem[Xin et~al.(2024)Xin, Luo, Zhou, Du, Liu, Fan, Li, and Du]{xin2024parameter}
Xin, Y.; Luo, S.; Zhou, H.; Du, J.; Liu, X.; Fan, Y.; Li, Q.; Du, Y.
\newblock Parameter-Efficient Fine-Tuning for Pre-Trained Vision Models: {A} Survey.
\newblock {\em CoRR} {\bf 2024}, {\em abs/2402.02242}.
\newblock {\url{https://doi.org/10.48550/arXiv.2402.02242}}.

\bibitem[Wang et~al.(2022{\natexlab{a}})Wang, Zhang, Lee, Zhang, Sun, Ren, Su, Perot, Dy, and Pfister]{wang2022learning}
Wang, Z.; Zhang, Z.; Lee, C.Y.; Zhang, H.; Sun, R.; Ren, X.; Su, G.; Perot, V.; Dy, J.; Pfister, T.
\newblock Learning to Prompt for Continual Learning.
\newblock In Proceedings of the 2022 IEEE/CVF Conference on Computer Vision and Pattern Recognition (CVPR),  2022, pp. 139--149.
\newblock {\url{https://doi.org/10.1109/CVPR52688.2022.00024}}.

\bibitem[Wang et~al.(2022{\natexlab{b}})Wang, Zhang, Ebrahimi, Sun, Zhang, Lee, Ren, Su, Perot, Dy, and Pfister]{wang2022dualprompt}
Wang, Z.; Zhang, Z.; Ebrahimi, S.; Sun, R.; Zhang, H.; Lee, C.Y.; Ren, X.; Su, G.; Perot, V.; Dy, J.;  et~al.
\newblock DualPrompt: Complementary Prompting for Rehearsal-Free Continual Learning.
\newblock In Proceedings of the Computer Vision -- ECCV 2022: 17th European Conference, Tel Aviv, Israel, October 23--27, 2022, Proceedings, Part XXVI. Springer-Verlag,  2022, pp. 631--648.
\newblock {\url{https://doi.org/10.1007/978-3-031-19809-0_36}}.

\bibitem[Gao et~al.(2023)Gao, Zhao, Sun, Xi, Zhang, Ghanem, and Zhang]{gao2023unified}
Gao, Q.; Zhao, C.; Sun, Y.; Xi, T.; Zhang, G.; Ghanem, B.; Zhang, J.
\newblock A Unified Continual Learning Framework with General Parameter-Efficient Tuning.
\newblock In Proceedings of the 2023 IEEE/CVF International Conference on Computer Vision (ICCV),  2023, pp. 11449--11459.
\newblock {\url{https://doi.org/10.1109/ICCV51070.2023.01055}}.

\bibitem[Zhou et~al.(2024)Zhou, Cai, Ye, Zhan, and Liu]{zhou2023revisitingclassincrementallearningpretrained}
Zhou, D.W.; Cai, Z.W.; Ye, H.J.; Zhan, D.C.; Liu, Z.
\newblock Revisiting Class-Incremental Learning with Pre-Trained Models: Generalizability and Adaptivity are All You Need.
\newblock {\em International Journal of Computer Vision} {\bf 2024}, {\em 133},~1012--1032.
\newblock {\url{https://doi.org/10.1007/s11263-024-02218-0}}.

\bibitem[Wu et~al.(2025)Wu, Piao, Huang, Wang, Li, Pfister, Meng, Ma, and Wei]{wu2025sdlora}
Wu, Y.; Piao, H.; Huang, L.K.; Wang, R.; Li, W.; Pfister, H.; Meng, D.; Ma, K.; Wei, Y.
\newblock SD-LoRA: Scalable Decoupled Low-Rank Adaptation for Class Incremental Learning.
\newblock In Proceedings of the The Thirteenth International Conference on Learning Representations, {ICLR} 2025,  2025.
\newblock {\url{https://doi.org/10.48550/arXiv.2501.13198}}.

\bibitem[Wang et~al.(2022)Wang, Huang, and Hong]{wang2022s}
Wang, Y.; Huang, Z.; Hong, X.
\newblock S-Prompts Learning with Pre-trained Transformers: An Occam's Razor for Domain Incremental Learning.
\newblock In Proceedings of the Advances in Neural Information Processing Systems 35,  2022.
\newblock {\url{https://doi.org/10.48550/arXiv.2207.12819}}.

\bibitem[Jacobs et~al.(1991)Jacobs, Jordan, Nowlan, and Hinton]{6797059}
Jacobs, R.A.; Jordan, M.I.; Nowlan, S.J.; Hinton, G.E.
\newblock Adaptive Mixtures of Local Experts.
\newblock {\em Neural Computation} {\bf 1991}, {\em 3},~79--87.
\newblock {\url{https://doi.org/10.1162/neco.1991.3.1.79}}.

\bibitem[Aljundi et~al.(2019)Aljundi, Kelchtermans, and Tuytelaars]{aljundi2019task}
Aljundi, R.; Kelchtermans, K.; Tuytelaars, T.
\newblock Task-Free Continual Learning.
\newblock In Proceedings of the 2019 IEEE/CVF Conference on Computer Vision and Pattern Recognition (CVPR),  2019, pp. 11246--11255.
\newblock {\url{https://doi.org/10.1109/CVPR.2019.01151}}.

\bibitem[Rebuffi et~al.(2017)Rebuffi, Kolesnikov, Sperl, and Lampert]{rebuffi2017icarl}
Rebuffi, S.A.; Kolesnikov, A.; Sperl, G.; Lampert, C.H.
\newblock iCaRL: Incremental Classifier and Representation Learning.
\newblock In Proceedings of the 2017 IEEE Conference on Computer Vision and Pattern Recognition (CVPR),  2017, pp. 5533--5542.
\newblock {\url{https://doi.org/10.1109/CVPR.2017.587}}.

\bibitem[Yu et~al.(2020)Yu, Twardowski, Liu, Herranz, Wang, Cheng, Jui, and van~de Weijer]{yu2020semantic}
Yu, L.; Twardowski, B.; Liu, X.; Herranz, L.; Wang, K.; Cheng, Y.; Jui, S.; van~de Weijer, J.
\newblock Semantic Drift Compensation for Class-Incremental Learning.
\newblock In Proceedings of the 2020 IEEE/CVF Conference on Computer Vision and Pattern Recognition (CVPR),  2020, pp. 6980--6989.
\newblock {\url{https://doi.org/10.1109/CVPR42600.2020.00701}}.

\bibitem[Wang et~al.(2022)Wang, Zhou, Ye, and Zhan]{wang2022foster}
Wang, F.; Zhou, D.; Ye, H.; Zhan, D.
\newblock {FOSTER:} Feature Boosting and Compression for Class-Incremental Learning.
\newblock In Proceedings of the Computer Vision - {ECCV} 2022 - 17th European Conference, Tel Aviv, Israel, October 23-27, 2022, Proceedings, Part {XXV}; Avidan, S.; Brostow, G.J.; Ciss{\'{e}}, M.; Farinella, G.M.; Hassner, T., Eds. Springer,  2022, Vol. 13685, {\em Lecture Notes in Computer Science}, pp. 398--414.
\newblock {\url{https://doi.org/10.1007/978-3-031-19806-9\_23}}.

\bibitem[Smith et~al.(2023)Smith, Karlinsky, Gutta, Cascante-Bonilla, Kim, Arbelle, Panda, Feris, and Kira]{smith2023coda}
Smith, J.S.; Karlinsky, L.; Gutta, V.; Cascante-Bonilla, P.; Kim, D.; Arbelle, A.; Panda, R.; Feris, R.; Kira, Z.
\newblock CODA-Prompt: COntinual Decomposed Attention-Based Prompting for Rehearsal-Free Continual Learning.
\newblock In Proceedings of the 2023 IEEE/CVF Conference on Computer Vision and Pattern Recognition (CVPR),  2023, pp. 11909--11919.
\newblock {\url{https://doi.org/10.1109/CVPR52729.2023.01146}}.

\bibitem[Yu et~al.(2024)Yu, Zhuge, Zhang, Hu, Wang, Lu, and He]{MoE-Adapters}
Yu, J.; Zhuge, Y.; Zhang, L.; Hu, P.; Wang, D.; Lu, H.; He, Y.
\newblock Boosting Continual Learning of Vision-Language Models via Mixture-of-Experts Adapters.
\newblock In Proceedings of the 2024 IEEE/CVF Conference on Computer Vision and Pattern Recognition (CVPR),  2024, pp. 23219--23230.
\newblock {\url{https://doi.org/10.1109/CVPR52733.2024.02191}}.

\bibitem[Riquelme et~al.(2021)Riquelme, Puigcerver, Mustafa, Neumann, Jenatton, Pinto, Keysers, and Houlsby]{riquelme2021scaling}
Riquelme, C.; Puigcerver, J.; Mustafa, B.; Neumann, M.; Jenatton, R.; Pinto, A.S.; Keysers, D.; Houlsby, N.
\newblock Scaling vision with sparse mixture of experts.
\newblock In Proceedings of the Proceedings of the 35th International Conference on Neural Information Processing Systems, Red Hook, NY, USA,  2021; NIPS '21.
\newblock {\url{https://doi.org/10.5555/3540261.3540918}}.

\bibitem[Gou et~al.(2023)Gou, Liu, Chen, Hong, Xu, Li, Yeung, Kwok, and Zhang]{gou2023mixture}
Gou, Y.; Liu, Z.; Chen, K.; Hong, L.; Xu, H.; Li, A.; Yeung, D.Y.; Kwok, J.T.; Zhang, Y.
\newblock Mixture of cluster-conditional lora experts for vision-language instruction tuning.
\newblock {\em arXiv preprint arXiv:2312.12379} {\bf 2023}.
\newblock {\url{https://doi.org/10.48550/arXiv.2312.12379}}.

\bibitem[Hu et~al.(2022)Hu, Shen, Wallis, Allen{-}Zhu, Li, Wang, Wang, and Chen]{hu2022lora}
Hu, E.J.; Shen, Y.; Wallis, P.; Allen{-}Zhu, Z.; Li, Y.; Wang, S.; Wang, L.; Chen, W.
\newblock LoRA: Low-Rank Adaptation of Large Language Models.
\newblock In Proceedings of the The Tenth International Conference on Learning Representations, {ICLR} 2022,  2022.
\newblock {\url{https://doi.org/10.48550/arXiv.2106.09685}}.

\bibitem[Jin et~al.(2025)Jin, Zhu, Yuan, and Yan]{jin2025moe}
Jin, P.; Zhu, B.; Yuan, L.; Yan, S.
\newblock MoE++: Accelerating Mixture-of-Experts Methods with Zero-Computation Experts.
\newblock In Proceedings of the The Thirteenth International Conference on Learning Representations, {ICLR} 2025, Singapore, April 24-28, 2025,  2025.

\bibitem[Dou et~al.(2023)Dou, Zhou, Liu, Gao, Zhao, Shen, Zhou, Xi, Wang, Fan, Pu, Zhu, Zheng, Gui, Zhang, and Huang]{dou2023loramoe}
Dou, S.; Zhou, E.; Liu, Y.; Gao, S.; Zhao, J.; Shen, W.; Zhou, Y.; Xi, Z.; Wang, X.; Fan, X.;  et~al.
\newblock LoRAMoE: Revolutionizing Mixture of Experts for Maintaining World Knowledge in Language Model Alignment.
\newblock {\em CoRR} {\bf 2023}, {\em abs/2312.09979},  \href{http://arxiv.org/abs/2312.09979}{{\normalfont [2312.09979]}}.
\newblock {\url{https://doi.org/10.48550/arXiv.2312.09979}}.

\bibitem[Wang et~al.(2023)Wang, Xie, Zhang, Huang, Su, and Zhu]{wang2023hierarchical}
Wang, L.; Xie, J.; Zhang, X.; Huang, M.; Su, H.; Zhu, J.
\newblock Hierarchical Decomposition of Prompt-Based Continual Learning: Rethinking Obscured Sub-optimality.
\newblock In Proceedings of the Advances in Neural Information Processing Systems 36: Annual Conference on Neural Information Processing Systems 2023, NeurIPS 2023, New Orleans, LA, USA, December 10 - 16, 2023; Oh, A.; Naumann, T.; Globerson, A.; Saenko, K.; Hardt, M.; Levine, S., Eds.,  2023.
\newblock {\url{https://doi.org/10.5555/3666122.3669144}}.

\bibitem[Dosovitskiy et~al.(2021)Dosovitskiy, Beyer, Kolesnikov, Weissenborn, Zhai, Unterthiner, Dehghani, Minderer, Heigold, Gelly, Uszkoreit, and Houlsby]{dosovitskiy2020image}
Dosovitskiy, A.; Beyer, L.; Kolesnikov, A.; Weissenborn, D.; Zhai, X.; Unterthiner, T.; Dehghani, M.; Minderer, M.; Heigold, G.; Gelly, S.;  et~al.
\newblock An Image is Worth 16x16 Words: Transformers for Image Recognition at Scale.
\newblock In Proceedings of the 9th International Conference on Learning Representations, {ICLR} 2021,  2021.
\newblock {\url{https://doi.org/10.48550/arXiv.2010.11929}}.

\bibitem[He et~al.(2015)He, Zhang, Ren, and Sun]{he2015delving}
He, K.; Zhang, X.; Ren, S.; Sun, J.
\newblock Delving Deep into Rectifiers: Surpassing Human-Level Performance on ImageNet Classification.
\newblock In Proceedings of the 2015 IEEE International Conference on Computer Vision (ICCV),  2015, pp. 1026--1034.
\newblock {\url{https://doi.org/10.1109/ICCV.2015.123}}.

\bibitem[Tao et~al.(2020)Tao, Hong, Chang, Dong, Wei, and Gong]{tao2020few}
Tao, X.; Hong, X.; Chang, X.; Dong, S.; Wei, X.; Gong, Y.
\newblock Few-Shot Class-Incremental Learning.
\newblock In Proceedings of the 2020 IEEE/CVF Conference on Computer Vision and Pattern Recognition (CVPR),  2020, pp. 12180--12189.
\newblock {\url{https://doi.org/10.1109/CVPR42600.2020.01220}}.

\bibitem[Zhai et~al.(2019)Zhai, Puigcerver, Kolesnikov, Ruyssen, Riquelme, Lucic, Djolonga, Pinto, Neumann, Dosovitskiy, et~al.]{zhai2019large}
Zhai, X.; Puigcerver, J.; Kolesnikov, A.; Ruyssen, P.; Riquelme, C.; Lucic, M.; Djolonga, J.; Pinto, A.S.; Neumann, M.; Dosovitskiy, A.;  et~al.
\newblock A large-scale study of representation learning with the visual task adaptation benchmark.
\newblock {\em arXiv preprint arXiv:1910.04867} {\bf 2019}.

\bibitem[Krizhevsky and Hinton(2009)]{krizhevsky2009learning}
Krizhevsky, A.; Hinton, G.
\newblock Learning multiple layers of features from tiny images.
\newblock Technical report, University of Toronto,  2009.

\bibitem[Wah et~al.(2011)Wah, Branson, Welinder, Perona, and Belongie]{wah2011caltech}
Wah, C.; Branson, S.; Welinder, P.; Perona, P.; Belongie, S.
\newblock The Caltech-UCSD Birds-200-2011 Dataset.
\newblock Technical report, California Institute of Technology,  2011.

\bibitem[Hendrycks et~al.(2021)Hendrycks, Basart, Mu, Kadavath, Wang, Dorundo, Desai, Zhu, Parajuli, Guo, et~al.]{hendrycks2021many}
Hendrycks, D.; Basart, S.; Mu, N.; Kadavath, S.; Wang, F.; Dorundo, E.; Desai, R.; Zhu, T.; Parajuli, S.; Guo, M.;  et~al.
\newblock The many faces of robustness: A critical analysis of out-of-distribution generalization.
\newblock In Proceedings of the Proceedings of the IEEE/CVF international conference on computer vision,  2021, pp. 8340--8349.

\bibitem[Zhang et~al.(2022)Zhang, Yin, Shao, and Liu]{zhang2022benchmarking}
Zhang, Y.; Yin, Z.; Shao, J.; Liu, Z.
\newblock Benchmarking omni-vision representation through the lens of visual realms.
\newblock In Proceedings of the European Conference on Computer Vision. Springer,  2022, pp. 594--611.

\bibitem[Park et~al.(2024)Park, Song, and Park]{park2024pre}
Park, K.H.; Song, K.; Park, G.M.
\newblock Pre-trained Vision and Language Transformers are Few-Shot Incremental Learners.
\newblock In Proceedings of the 2024 IEEE/CVF Conference on Computer Vision and Pattern Recognition (CVPR),  2024, pp. 23881--23890.
\newblock {\url{https://doi.org/10.1109/CVPR52733.2024.02254}}.

\bibitem[Liu et~al.(2024)Liu, Wang, Xiong, Chen, Wu, Guo, and Huang]{liu2024few}
Liu, C.; Wang, Z.; Xiong, T.; Chen, R.; Wu, Y.; Guo, J.; Huang, H.
\newblock Few-Shot Class Incremental Learning with Attention-Aware Self-adaptive Prompt.
\newblock In Proceedings of the Computer Vision - {ECCV} 2024 - 18th European Conference, Milan, Italy, September 29-October 4, 2024, Proceedings, Part {LXXXI}, Cham,  2024; pp. 1--18.
\newblock {\url{https://doi.org/10.1007/978-3-031-73004-7_1}}.

\bibitem[Russakovsky et~al.(2015)Russakovsky, Deng, Su, Krause, Satheesh, Ma, Huang, Karpathy, Khosla, Bernstein, Berg, and Fei{-}Fei]{russakovsky2015imagenet}
Russakovsky, O.; Deng, J.; Su, H.; Krause, J.; Satheesh, S.; Ma, S.; Huang, Z.; Karpathy, A.; Khosla, A.; Bernstein, M.S.;  et~al.
\newblock ImageNet Large Scale Visual Recognition Challenge.
\newblock {\em Int. J. Comput. Vis.} {\bf 2015}, {\em 115},~211--252.
\newblock {\url{https://doi.org/10.1007/S11263-015-0816-Y}}.

\bibitem[Zhu et~al.(2025)Zhu, Zhang, Dong, and Koniusz]{zhu2025bilora}
Zhu, H.; Zhang, Y.; Dong, J.; Koniusz, P.
\newblock BiLoRA: Almost-Orthogonal Parameter Spaces for Continual Learning.
\newblock In Proceedings of the Proceedings of the IEEE/CVF Conference on Computer Vision and Pattern Recognition (CVPR),  June 2025, pp. 25613--25622.

\bibitem[Zhou et~al.(2024)Zhou, Sun, Ye, and Zhan]{zhou2024ease}
Zhou, D.W.; Sun, H.L.; Ye, H.J.; Zhan, D.C.
\newblock Expandable Subspace Ensemble for Pre-Trained Model-Based Class-Incremental Learning.
\newblock In Proceedings of the 2024 IEEE/CVF Conference on Computer Vision and Pattern Recognition (CVPR),  2024, pp. 23554--23564.
\newblock {\url{https://doi.org/10.1109/CVPR52733.2024.02223}}.

\bibitem[D'Alessandro et~al.(2023)D'Alessandro, Alonso, Calabrés, and Galar]{10350931}
D'Alessandro, M.; Alonso, A.; Calabrés, E.; Galar, M.
\newblock Multimodal Parameter-Efficient Few-Shot Class Incremental Learning.
\newblock In Proceedings of the 2023 IEEE/CVF International Conference on Computer Vision Workshops (ICCVW),  2023, pp. 3385--3395.
\newblock {\url{https://doi.org/10.1109/ICCVW60793.2023.00364}}.

\bibitem[Kim and Han(2023)]{kim2023stabilityplasticity}
Kim, D.; Han, B.
\newblock On the Stability-Plasticity Dilemma of Class-Incremental Learning.
\newblock In Proceedings of the Proceedings of the IEEE/CVF Conference on Computer Vision and Pattern Recognition (CVPR),  2023, pp. 20196--20204.

\bibitem[Caron et~al.(2021)Caron, Touvron, Misra, Jegou, Mairal, Bojanowski, and Joulin]{caron2021emerging}
Caron, M.; Touvron, H.; Misra, I.; Jegou, H.; Mairal, J.; Bojanowski, P.; Joulin, A.
\newblock Emerging Properties in Self-Supervised Vision Transformers.
\newblock In Proceedings of the 2021 IEEE/CVF International Conference on Computer Vision (ICCV),  2021, pp. 9630--9640.
\newblock {\url{https://doi.org/10.1109/ICCV48922.2021.00951}}.

\bibitem[Chen et~al.(2022)Chen, Hsieh, and Gong]{chen2021vision}
Chen, X.; Hsieh, C.; Gong, B.
\newblock When Vision Transformers Outperform ResNets without Pre-training or Strong Data Augmentations.
\newblock In Proceedings of the The Tenth International Conference on Learning Representations, {ICLR} 2022,  2022.
\newblock {\url{https://doi.org/10.48550/arXiv.2106.01548}}.

\end{thebibliography}
\end{document}